\documentclass[manuscript,screen]{acmart}

\AtBeginDocument{%
  \providecommand\BibTeX{{%
    \normalfont B\kern-0.5em{\scshape i\kern-0.25em b}\kern-0.8em\TeX}}}

\setcopyright{acmcopyright}
\copyrightyear{2018}
\acmYear{2018}
\acmDOI{XXXXXXX.XXXXXXX}

\usepackage{amsmath}
\usepackage{booktabs}
\usepackage[ruled,vlined]{algorithm2e}
\usepackage{multirow}
\usepackage{caption}
\usepackage{subcaption}
\usepackage{xcolor}
\begin{document}
%
\title{A Survey for Biomedical Text Summarization: From Pre-trained to Large Language Models}

\author{Qianqian Xie}
\affiliation{%
  \institution{Department
of Computer Science, University of Manchester}
  \city{Manchester}
  \country{United Kingdom}
  }
\email{xqq.sincere@gmail.com}
\author{Zheheng Luo}
\affiliation{%
  \institution{Department
of Computer Science, University of Manchester}
  \city{Manchester}
  \country{United Kingdom}}
\email{zheheng.luo@postgrad.manchester}
\author{Benyou Wang}
\affiliation{%
  \institution{School of Data Science, The Chinese University of Hong Kong}
  \city{Shengzhen}
  \country{China}}
\email{wangbenyou@cuhk.edu.cn}
\author{Sophia Ananiadou}
\affiliation{%
  \institution{Department
of Computer Science, University of Manchester}
  \city{Manchester}
  \country{United Kingdom}
  }
\email{Sophia.Ananiadou@manchester.ac.uk}

\renewcommand{\shortauthors}{Qianqian and Zheheng, et al.}

\begin{abstract}
The exponential growth of biomedical texts such as biomedical literature and electronic health records (EHRs), poses a significant challenge for clinicians and researchers to access clinical information efficiently.
To tackle this challenge, biomedical text summarization (BTS) has been proposed as a solution to support clinical information retrieval and management.
BTS aims at generating concise summaries that distill key information from single or multiple biomedical documents. 
In recent years, the rapid advancement of fundamental natural language processing (NLP) techniques, from pre-trained language models (PLMs) to large language models (LLMs), has greatly facilitated the progress of BTS.
This growth has led to numerous proposed summarization methods, datasets, and evaluation metrics, raising the need for a comprehensive and up-to-date survey for BTS. 
In this paper, we present a systematic review of recent advancements in BTS, leveraging cutting-edge NLP techniques from PLMs to LLMs, to help understand the latest progress, challenges, and future directions. 
We begin by introducing the foundational concepts of BTS, PLMs and LLMs, followed by an in-depth review of available datasets, recent approaches, and evaluation metrics in BTS. We finally discuss existing challenges and promising future directions in the era of LLMs.
To facilitate the research community, we line up open resources including available datasets, recent approaches, codes, evaluation metrics, and the leaderboard in a public project: \url{https://github.com/KenZLuo/Biomedical-Text-Summarization-Survey/tree/master}.
We believe that this survey will be a useful resource to researchers, allowing them to quickly track recent advancements and provide guidelines for future BTS research within the research community.
\end{abstract}



\maketitle

\section{Introduction}\label{sec:introduction}

The rapidly increasing of unstructured clinical text information, such as biomedical literature~\cite{lu2011pubmed} and clinical notes~\cite{jensen2012mining}, presents a significant challenge for researchers and clinicians seeking effective access to required information.
To tackle the challenge, the field of biomedical text summarization (BTS)~\cite{maybury1999advances,xie2023factreranker} has emerged to assist users in seeking information efficiently.
It aims to shorten single or multiple lengthy biomedical documents into condensed summaries that capture the most important information~\cite{mishra2014text}.
By reading these summaries, researchers and clinicians can quickly grasp the main ideas of extensive biomedical texts, saving time and effort.
The BTS task has diverse real-world applications including but not limited to aiding evidence-based medicine~\cite{przybyla2018prioritising}, clinical information management~\cite{esteva2021covid}, and clinical decision support~\cite{demner2009can}.

There has been a remarkable advancement in the field of BST in recent years, due to the evolution of fundamental NLP techniques ranging from pre-trained language models (PLMs)~\cite{kenton2019bert} to large language models (LLMs)~\cite{brown2020language}.
PLMs, namely language models pre-trained with the large scale of unlabeled data in a self-supervised manner, possess the ability to capture the common sense and lexical knowledge inherited in the training data~\cite{roberts2020much}. 
With the acquired knowledge, PLMs have significantly enhanced a wide range of NLP tasks with the process of fine-tuning.
Researchers further enhanced the efficiency of PLMs by increasing the size of models and training datasets, resulting in the emergence of large language models (LLMs)~\cite{brown2020language,ouyang2022training}.
Compared with PLMs, LLMs possess a remarkable ability for natural language understanding and generation.
Moreover, LLMs possess the emergent ability of in-context learning, allowing them to perform diverse tasks without the need for supervised training, simply by following natural language instructions~\cite{wei2023larger}.
For example, GPT-4~\cite{bubeck2023sparks}, an LLM released by OpenAI, has shown human-level capabilities in the zero-shot setting across various domains, including vision, coding, mathematics, medicine et al.
This growth has revolutionized NLP and also led to numerous proposed summarization methods, datasets, and evaluation metrics for BTS~\cite{wallace2021generating,luo2022readability,guo2021automated}. 
In these methods, general-domain PLMs such as BERT~\cite{kenton2019bert} and BART~\cite{lewis2020bart}, or domain-specific PLMs such as BioBERT~\cite{lee2020biobert} are employed as the backbone model for encoding input texts and then fine-tuned with the unstructured biomedical dataset of the BTS task.
It allows the semantic knowledge captured in PLMs to be transferred to the BTS task, resulting in more conclusive and informative summaries. 
Latest, there is an increasing number of studies exploring LLMs for zero-shot summary generation and evaluation of biomedical texts~\cite{Xie2023.04.18.23288752,shaib2023summarizing}, where LLMs have shown comparable or even better performance than PLMs without any supervised training.

These advances raise the need for a comprehensive and up-to-date survey for BTS. 
Although the availability of previous surveys for BTS with traditional machine learning and deep learning techniques~\cite{afantenos2005summarization,mishra2014text,pivovarov2015automated,wang2021systematic}, 
there exists a literature gap concerning recent developments in BTS based on PLMs and LLMs. 
This paper aims to bridge this gap by conducting an in-depth survey of the recent work exploring PLMs and LLMs for BTS. 
We conduct a systematic review of benchmark datasets, summarization approaches, and evaluation methods of the task. 
We categorize and delve into a detailed discussion of existing methods according to their leveraging of PLMs and LLMs, and analyze their limitations and outlook for future directions.
We hope this paper can be a timely survey for researchers in the research community to quickly track recent progress, challenges, and promising future directions. 

\textbf{Compared with existing surveys} Afantenos et al~\cite{afantenos2005summarization} was the earliest survey that summarized traditional NLP and machine learning methods for medical document summarization.
Mishra et al~\cite{mishra2014text} reviewed text summarization methods for biomedical literature and electronic health records (EHRs), between January 2000 and October 2013. 
Pivovarov et al~\cite{pivovarov2015automated} examined automated summarization methods for electronic health records. 
Most methods summarized in these surveys are traditional machine learning methods based on feature engineering.
With the prosperity of deep learning since 2014, deep neural networks became the mainstream method for biomedical text summarization.
Recently, Wang et al~\cite{wang2021systematic} investigated deep learning-based text summarization approaches for both biomedical literature and EHRs between January 2013 to April 2021.
Nevertheless, recent research with PLMs and LLMs for biomedical text summarization was not included in this survey.
Wang et al~\cite{wang2021pre} summarized PLMs-based methods for biomedical NLP, and briefly introduced PLMs for the biomedical text summarization task as one of the various tasks.
Compared with them, we provide a more comprehensive and focused overview for PLMs and LLMs on biomedical text summarization including benchmark datasets, evaluation metrics, and limitations et al. 
To the best of our knowledge, our paper is the first review that surveys recent LLMs and PLMs-based methods for BTS. 

\textbf{Paper collection}
We collect representative works since 2018 that are published in conferences and journals of computer science and biomedical science such as ACL, EMNLP, COLING, NAACL, AAAI, Bioinformatics, BioNLP, JAMIA, AMIA, NPJ digital medicine et al. 
We use PubMed, google scholar, ACM Digital Library, and IEEE Xplore Digital Library as the search engine and the database. 
We search with keywords including "biomedical/medical/clinical summarization", "medical dialogue summarization", "radiology report summarization", "large language models", "pre-trained language models", "medical/biomedical/clinical evaluation", "medical/biomedical/clinical datasets".

\begin{figure*}[t!]
    \centering
    \includegraphics[width=1.0\linewidth]{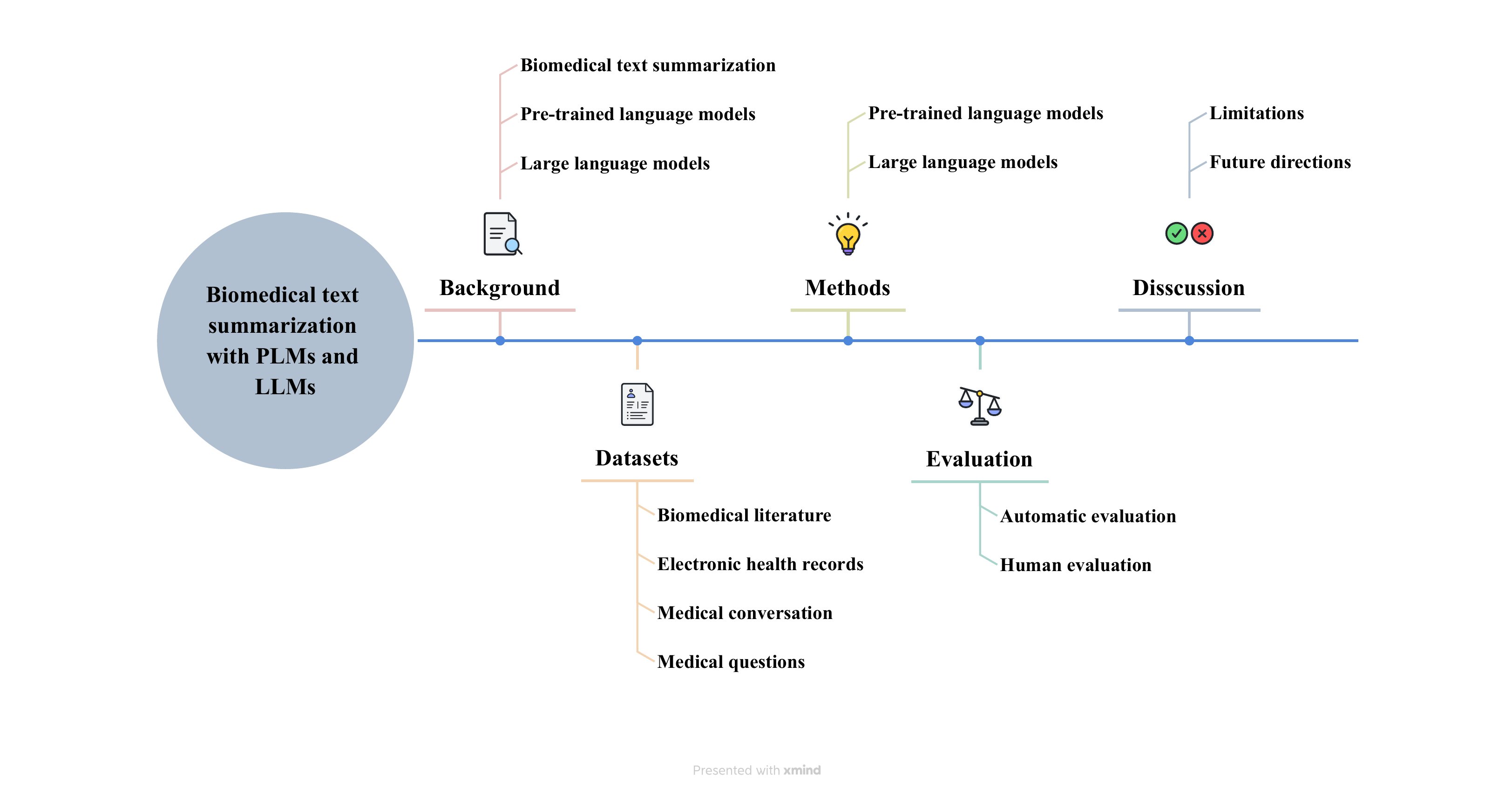}
    \caption{Overview of biomedical text summarization with PLMs and LLMs.}
    \label{fig:0}
\end{figure*}

\textbf{Organization of the paper} We will first introduce the background of biomedical text summarization, pre-trained language models, and large language models in Section \ref{sec:back}.
Then Section \ref{sec:data} will describe existing benchmark datasets.
Representative methods with PLMs and LLMs will be categorized and discussed in Section \ref{sec:method}.
We introduce evaluation methods in Section \ref{sec:eval}.
We next discuss limitations and future directions in Section \ref{sec:lim}. 
Finally, we make a conclusion in Section \ref{sec:con}.
Figure \ref{fig:0} shows the proposed overview of biomedical text summarization with pre-trained language models and large language models.
\begin{figure*}
    \centering
    \includegraphics[width=1\linewidth]{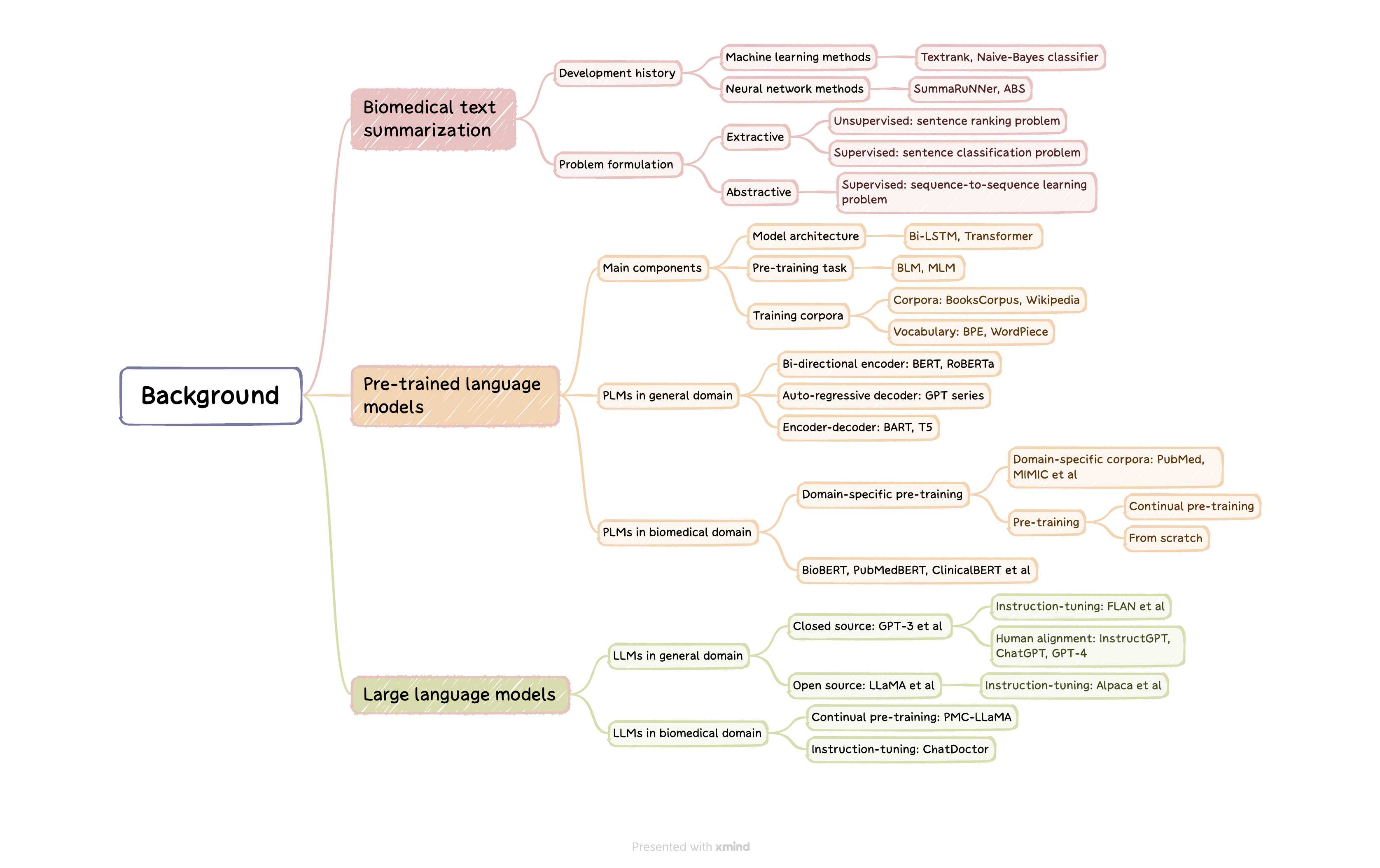}
    \caption{Overview of background.}
    \label{fig:1}
\end{figure*}
\begin{figure*}[t!]
    \centering
    \includegraphics[width=0.8\linewidth]{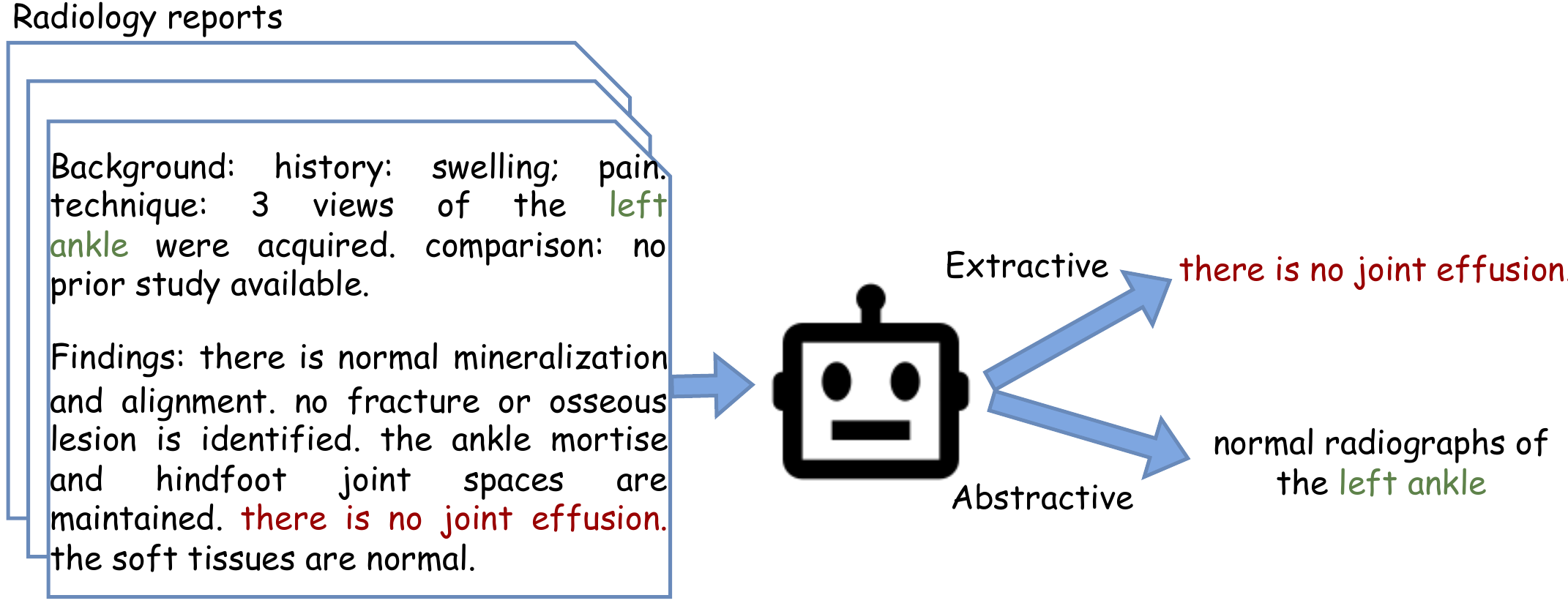}
    \caption{The example of extractive and abstractive biomedical text summarization.}
    \label{fig:e}
\end{figure*}

\section{Background}
\label{sec:back}
In this section, we first review biomedical text summarization, pre-trained language models, and large language models, which are essential concepts used in this survey.
The overview of the background section is shown in Figure \ref{fig:1}.
\subsection{Biomedical Text Summarization}
Biomedical text summarization~\cite{pivovarov2015automated,wang2021systematic} aims to shorten single or multiple biomedical documents into a condensed summary that preserves the most important information.
In general, automated summarization approaches are divided into extractive summarization methods~\cite{nallapati2017summarunner} and abstractive summarization methods~\cite{lin2019abstractive} according to the output of summaries as shown in Figure \ref{fig:e}.
Extractive methods select key sentences from original documents and concatenate them into a summary, while abstractive methods generate new sentences as the summary based on the original documents.
Compared with extractive summarization, 
abstractive summarization presents a more significant challenge since it involves generating informative sentences from a large vocabulary, syntactic adjustment, and paraphrasing, to produce factually consistent summaries.
Formally, let's assume $C$ as a biomedical corpus with $D$ documents, $d \in C$ is a document consisting of $m$ sentences: $d=\{s_{1},\cdots,s_{m}\}$. 
We also assume the gold summary of the document $d$ as $t_d$.
For biomedical scientific papers, abstract of papers are generally deemed as their gold summaries.

Automatic biomedical summarization methods are largely facilitated and inspired by NLP methods in the general domain.
The earliest methods are traditional machine learning methods such as Naive-Bayes classifier~\cite{kupiec1995trainable} and graph-based ranking methods such as TextRank~\cite{mihalcea2004textrank}.
With the prosperity of deep learning since 2014, neural network methods~\cite{lecun2015deep} have been the mainstream method for both extractive and abstractive summarization of biomedical texts.
Neural extractive methods~\cite{nallapati2017summarunner} formulate the extractive task as the binary classification problem.
This involves predicting the labels (1 or 0) of sentences in original documents to select the sentences to be included in the summary.
Neural abstractive methods~\cite{rush2015neural} model the abstractive task as the text generation problem that generates new sentences based on the sequence-to-sequence~\cite{sutskever2014sequence} framework.

\textbf{Extractive summarization} 
For a document $d$ with $m$ sentences, extractive summarization methods aim to select a subset of $o$ sentences from $d$, $o \ll m$.
Neural extractive methods can be classified into unsupervised methods and supervised methods.
The unsupervised methods~\cite{mihalcea2004textrank,erkan2004lexrank} model the extractive task into the sentence ranking problem.
They usually represent sentences with advanced representation techniques such as word embeddings, and use the unsupervised ranking method to select important sentences based on their representations.
Supervised methods~\cite{nallapati2017summarunner}, adopt human-written abstractive summaries as the gold summaries whose sentences are not in the original documents.
To train the extractive models, they are first required to generate binary labels for sentences according to the gold summaries.
To this end, they generally adopt unsupervised sentence selection methods such as the greedy search algorithm to generate the oracle summary for each document with sentences that are most semantically similar to the gold summary.
Therefore, sentences that are included in the oracle summary are labeled with $1$, while the remaining sentences are labeled with $0$.

Most supervised neural extractive methods consist of the neural network-based encoder and classifier.
The neural network-based encoder is used to capture the contextual information of input documents and generate vector representations of sentences.
The classifier is to predict labels of sentences according to their vector representations.
The objective is to maximize the log-likelihood of the observed labels of sentences:
\begin{equation}
    \log p(y|C;\theta) = \sum_{d \in C}\sum_{i=1}^m \log p(y_i^d|d;\theta)
\end{equation}
where $y^d_i$ is the ground truth label of sentence $s_i$ in document $d$, $\theta$ is the parameter set of the model.

\textbf{Abstractive summarization} Neural abstractive methods build the abstractive task as the sequence-to-sequence learning problem.
Most of them utilize the encoder-decoder framework~\cite{sutskever2014sequence}, which consists of the neural network-based encoder and decoder.
Similar to extractive methods, the encoder is used to yield vector representations of input documents.
The decoder is to generate the target summary sequentially with representations from the encoder.
The model is optimized with the objective to maximize the log-likelihood of target words in the gold summary. 
\begin{equation}
    \log p(t|C;\theta) = \sum_{d \in C}\sum_{i=1}^n \log p(t_i^d|d;\theta)
\end{equation}
where $t^d_i$ is the $i$-th word in the gold summary $t^d$ of the document $d$, $n \ll m$.

In recent years, PLMs have become the dominant method of biomedical text summarization.
PLMs-based methods~\cite{du2020biomedical,xie2022pre,luo2023citationsum,luo2022readability,xie2023factreranker} have a similar framework to neural methods for extractive and abstractive tasks, while PLMs are more powerful than neural networks in encoding biomedical texts.
We will next introduce the pre-trained language models.

\subsection{Pre-trained Language Models}
Language model pre-training~\cite{bengio2000neural,mikolov2013distributed} has long been an active research area with the aim of learning low-dimensional vector representations from natural language, which are applicable and generalizable for downstream tasks.
The earliest unidirectional neural language models such as word2vec~\cite{mikolov2013distributed} and glove~\cite{pennington2014glove}, learn meaningful word embeddings via estimating the probability of the next word with the sequence of history words.
The bidirectional language models such as ELMo~\cite{peters-etal-2018-deep} are then proposed to further consider the bidirectional context of words. 
Bidirectional Encoder Representations from Transformers (BERT)~\cite{kenton2019bert} is the breakthrough work that advances the state-of-art of various NLP tasks.
BERT and its variants generally consist of two steps: pre-training and fine-tuning.
It proposes to first pre-train the deep models based on basic neural network structure such as transformer~\cite{vaswani2017attention} on the large scale of unlabeled data with a self-supervised learning task. 
Then the pre-trained parameters of deep models and task-specific parameters are fine-tuned on labeled data with downstream tasks. 
To provide a more comprehensive understanding of PLMs, we will further illustrate their core components.
For more details of PLMs, one can check the review~\cite{qiu2020pre}.

\textbf{Model architecture} 
The early language models such as ELMo and its predecessors~\cite{peters2017semi,peters-etal-2018-deep}, generally utilize Bi-LSTM~\cite{hochreiter1997long} as the backbone network structure, to capture bi-directional contextual information of texts.
However, Bi-LSTM has the limitation of parallelization and sequential computation with the growth of sequence length.
One breakthrough work is Transformer~\cite{vaswani2017attention}, with the self-attention-based model architecture, enabling it to handle parallel computation and model long-range dependencies of sequences efficiently.
It follows the encoder-decoder architecture with stacked multi-head self-attention and a point-wise fully connected feed-forward network.
Following the success of the Transformer model, most pre-trained language models have adopted this architecture, and use deep network architecture.
For example, the base model of BERT has 12 Transformer layers with the hidden size 768 and 12 self-attention heads. 

\begin{figure*}
    \centering
    \subfloat[Bi-directional encoder model.]{
   \begin{minipage}{.25\textwidth}
  \centering
  \includegraphics[width=.9\linewidth]{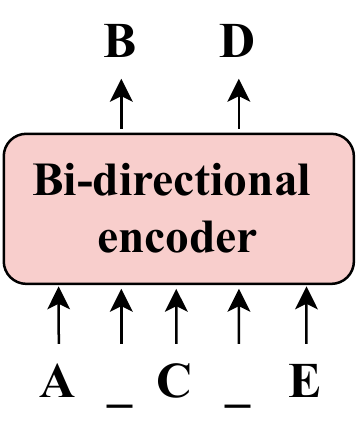}
\end{minipage}%
}
\subfloat[Autoregressive decoder model.]{
\begin{minipage}{.25\textwidth}
  \centering
  \includegraphics[width=.9\linewidth]{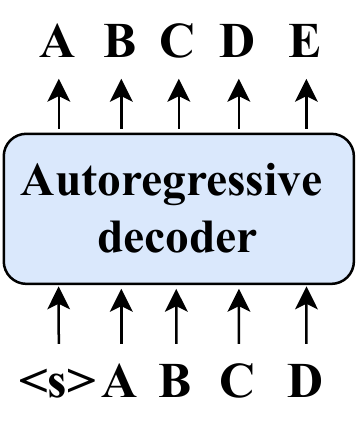}
\end{minipage}
}
\subfloat[Encoder-decoder models.]{
\begin{minipage}{.5\textwidth}
  \centering
  \includegraphics[width=.9\linewidth]{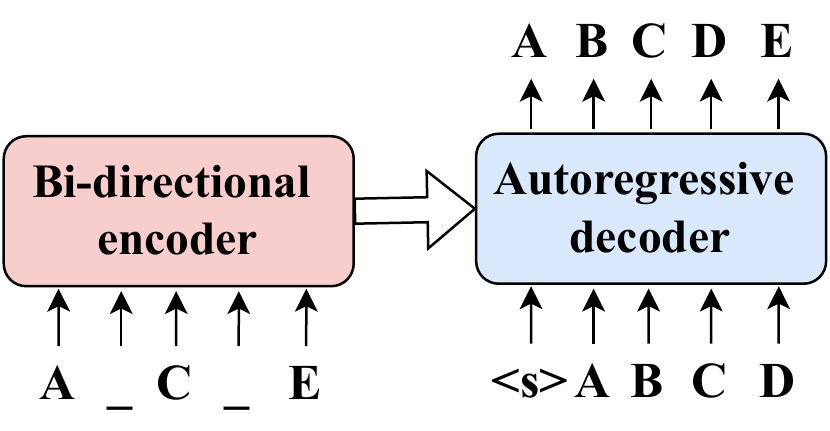}
\end{minipage}
}
\caption{Comparison of different language models.}
\label{fig:differ}
\end{figure*}

According to different model architectures, existing PLMs can be categorized into three types: bi-directional encoder language models, auto-regressive decoder language models, and encoder-decoder language models, as shown in Figure \ref{fig:differ}.
The bi-directional language models, i.e, BERT and its variants such as Roberta~\cite{liu2019roberta}, use only the encoder portion of Transformer without using the decoder structure in Transformer and predict masked tokens independently with the masked input sequence. 
However, they are hard to be used for generation tasks due to the absence of a decoder structure.
The auto-regressive language models such as GPT series~\cite{radford2018improving,radford2019language,brown2020language}, only use the decoder portion of the Transformer. 
They are trained to predict tokens in an auto-regressive manner, making them suitable for text-generation tasks.
However, they only consider leftward tokens during prediction and can only encode input tokens unidirectionally.
In contrast, encoder-decoder language models such as BART~\cite{lewis2020bart} use the full encoder and decoder architecture within Transformer. 
They incorporate the bi-directional encoding of masked input tokens and offer enhanced flexibility since there is no strict need to align the input tokens of the encoder and output tokens in the decoder.

\textbf{Training corpora} 
The majority of pre-trained language models (PLMs) utilize corpora from the general domain, such as BooksCorpus~\cite{zhu2015aligning} and Wikipedia. However, these models often encounter out-of-vocabulary (OOV) words that are not present in their vocabulary. To tackle this issue, PLMs employ sub-word tokenization techniques such as Byte-Pair Encoding (BPE)~\cite{sennrich2016neural} or WordPiece~\cite{kudo2018sentencepiece} to construct their vocabularies.

\textbf{Pre-training}
The pre-training on a large scale of unlabeled data is the key for language models to learn useful representations and parameters, which can be fine-tuned to downstream tasks.
The pre-training task of most previous language models follows the unidirectional language model~\cite{bengio2000neural}.
It aims to maximize the log-likelihood of words conditionally on history words:
\begin{equation}
    \mathcal{L}_{lm} = -\sum_{t=1}^T \log p(x_t|x_1,x_2,\cdots,x_{t-1})
\end{equation}
where $X=\{x_1,\cdots,x_{T}\}$ is a given text sequence with $T$ words.
The bidirectional language model is further proposed to capture contextual information of text from both directions. 
It combines both the left-to-right language model and the right-to-left language model:
\begin{equation}
\begin{split}
    \mathcal{L}_{blm} = &-\sum_{t=1}^T (\log p(x_t|x_1,x_2,\cdots,x_{t-1})\\
   &+ \log p(x_t|x_{t+1},x_{t+1},\cdots,x_T))
\end{split}
\end{equation}
Different from the bidirectional language model, PLMs such as BERT utilize the masked language model (MLM), which allows bi-directional self-supervised pre-training more efficiently.
It randomly selects 15\% tokens of the input text to be predicted, in which 80\% of them are replaced with the special token "[MASK]", and 10\% of them are replaced with other words in the vocabulary.
The objective is to maximize the log-likelihood of ground-truth words in the selected positions with masked text sequence:
\begin{equation}
    \mathcal{L}_{mlm} = -\sum_{\hat{x} \in m(x)} \log p(\hat{x}|X_m)
\end{equation}
where $X_m$ is the masked text sequence and $m(x)$ is the set of masked words.
Pre-training with MLM guides the language model to fully capture the contextual information embedded in the token sequence and generate more expressive representations on different levels such as tokens and sentences.

\textbf{Supervised Fine-tuning}
Self-supervised pre-training on large-scale corpora enables language models to acquire common sense knowledge and linguistic understanding in their pre-trained parameters and contextual representations. 
However, to effectively apply these models to specific downstream tasks, it is crucial to adapt them and their generated representations through supervised fine-tuning with task-specific objectives and datasets.
In the context of downstream tasks, task-specific inputs are initially fed into PLMs to generate contextual representations.
It requires choosing contextual representations on different levels and different task-specific layers stacked on top of language models, for corresponding tasks.
For instance, in the case of the extractive summarization task, previous research has commonly incorporated an additional classification layer to predict sentence labels based on the sentence-level representations generated by PLMs. 
By optimizing the model using the classification loss, both the parameters of PLMs and the task-specific parameters are refined.
Through this fine-tuning process, the linguistic and commonsense knowledge present in PLMs is effectively transferred into task-specific representations.
Thus, PLMs have significantly improved the performance in various NLP tasks.

\textbf{Biomedical Language Models}
Inspired by the great success of PLMs on NLP tasks, much attention has been devoted to applying PLMs to the biomedical domain including the task of biomedical text summarization.
However, most PLMs, including BERT, variants of BERT, T5~\cite{raffel2020exploring} et al, are pre-trained on texts of the general domain~\cite{zhu2015aligning}.
This poses a significant challenge when directly applying these models to biomedical texts.
The greatest issue is the terminologies and compound words in biomedical texts, most of which have never been mentioned in the general domain texts, and thus can't be covered by the vocabulary of PLMs.

To fill the gap, PLMs for the biomedical domain such as BioBERT~\cite{lee2020biobert}, BlueBERT~\cite{peng2019transfer}, and ClinicalBERT~\cite{alsentzer2019publicly} et al, have been proposed to further pre-train PLMs in the general domain with biomedical texts.
BioBERT~\cite{lee2020biobert} is the first biomedical language model that further pre-trains BERT on biomedical scientific texts including PubMed abstracts and PubMed Central full-text articles (PMC).
Following it, BlueBERT~\cite{peng2019transfer} conducts continual pre-training on clinical text corpus MIMIC-III~\cite{johnson2016mimic} along with biomedical scientific texts.
Yet they still adopt the same vocabulary as BERT, which limits their ability in encoding biomedical texts.
Different from them, SciBERT~\cite{beltagy2019scibert} builds a domain-specific vocabulary from scratch and conducts pre-training on scientific literature, in which 12\% articles are from the computer science domain and 82\% articles are from the biomedical domain.
PubMedBERT~\cite{gu2021domain} pre-trains their models with scientific papers solely in the biomedical domain.
Compared with PLMs in the general domain, biomedical PLMs can provide better representations that capture the contextual information of both normal tokens and domain-specific terms.
One can check the survey paper~\cite{wang2021pre} for more details on pre-trained language models for the biomedical domain.

\subsection{Large Language Models}
To improve the efficacy of language models, researchers have dedicated to scaling up language models~\cite{brown2020language,ouyang2022training}, by increasing the number of model parameters, expanding the size of training datasets, and enhancing computational resources. 
These lead to the emergence of large language models, namely transformer-based language models with billions of parameters and pre-trained with large amounts of text data.
LLMs usually have the same neural network architecture and pre-training task as PLMs introduced above.
The first representative LLM is GPT-3~\cite{brown2020language}, proposed by OpenAI in 2020.
It is an autoregressive decoder-based language model with parameters up to 175B, which only uses the decoder part of the transformer.
Subsequently, most LLMs use the autoregressive decoder of the transformer as the model architecture.
With the increased scale, LLMs have demonstrated remarkable advancements in natural language understanding, generation, and reasoning compared to smaller PLMs. 
They exhibit significantly enhanced abilities in handling complex language tasks.
Moreover, researchers have found that LLMs have the emergent ability (the ability not presented in small PLMs but LLMs~\cite{wei2022emergent}) of in-context learning, namely the capability to perform unseen tasks without any specific training data or with only a few training samples, by following natural language instructions.
Based on representative LLMs such as GPT-3~\cite{raffel2020exploring}, recent efforts have been explored in the following directions to improve the capability of LLMs:
\begin{itemize}
\item \textbf{Instruction Tuning} To enhance the generalization and in-context learning capabilities of LLMs, researchers such as T0~\cite{sanh2022multitask} and FLAN~\cite{wei2021finetuned}, further explored fine-tuning LLMs with diverse tasks expressed by natural language instruction, namely instruction tuning. 
The instruction tuning, makes LLMs to better understand and respond to instructions, enabling them to generalize more effectively to new tasks. 
During the instruction tuning process, researchers~\cite{chung2022scaling} have found certain factors including scaling the number of tasks, and model size and using chain-of-thought tuning data, that contribute to improving LLMs' generalization ability and reasoning ability on complex tasks.

\item  \textbf{Human Alignment} In addition to the capacity to follow instructions, it's crucial to ensure that large language models (LLMs) are guided by human preferences, avoiding generating unfaithful and toxic information~\cite{stiennon2020learning}. 
To achieve this, studies such as InstructGPT~\cite{ouyang2022training}, have explored fine-tuning LLMs with reinforcement learning from human feedback (RLHF)~\cite{christiano2017deep}, where the human preferences are used as the reward signal to fine-tune LLMs.
RLHF has demonstrated effectiveness in improving the truthfulness and toxicity of LLMs. ChatGPT\footnote{\url{https://openai.com/blog/chatgpt}} and GPT-4~\cite{openai2023gpt4} are two representative LLMs trained with RLHF by OpenAI, 
They have shown remarkable abilities in communicating with humans and dealing with complex tasks in diverse domains such as coding, vision, medicine, and law.
\item \textbf{Open Source} Due to the high cost of pre-training LLMs, most representative LLMs such as GPT-3, InstructGPT, ChatGPT, and GPT-4 are developed by technology companies with enough computation resources for commercial use. They are not openly accessible for public research or development. In response to this, efforts have been made to introduce openly available LLMs to aid future research. Notable examples include LLaMA~\cite{touvron2023llama}, a collection of efficient LLMs with parameters from 7B to 65B, RedPajama~\cite{together2023redpajama} with 7B parameters, and Falcon\footnote{\url{https://huggingface.co/blog/falcon}} with 40B parameters.
Despite having significantly fewer parameters, these openly accessible LLMs have demonstrated performance on par with larger models like GPT-3, which has 175B parameters, achieved through pre-training on extensive and high-quality text data.
\end{itemize}

\textbf{LLMs in Biomedical Domain}
LLMs have shown significant potential in the biomedical domain~\cite{nori2023capabilities}, with applications such as diagnostic assistance and drug discovery.
For example, GPT-4, one of the prominent LLMs, has outperformed the passing score of the United States Medical Licensing Examination (USMLE) by 20 points. 
Studies have been proposed to tailor LLMs to the biomedical domain by instruction tuning~\cite{singhal2022large} or continual pre-training with biomedical text data~\cite{wu2023pmc}.
While these techniques have further enhanced the performance of LLMs in medical tasks, all studies indicate that current LLMs are not yet prepared for reliable application in the medical field, due to their limitations in providing accurate, factual, safe, and unbiased information~\cite{xie2023faithful}. 

\section{Datasets}
\label{sec:data}
Unstructured biomedical texts used in text summarization methods involve various types, including biomedical literature, electronic health records (EHRs), medical conversations, and medical questions.
Details of these datasets are summarized in Table \ref{tab:bio}.
\begin{table*}[t]
    \begin{tabular}{c|cccc}
    \toprule
    \bf{Dataset}& \bf{Category}&\bf{Size}&\bf{Content}& \bf{Summarization Task}\\
    \midrule
    PubMed~\cite{cohan2018discourse} & Biomedical literature&133,215 &Full contents of articles&Single\\
SumPubMed~\cite{gupta2021sumpubmed}&Biomedical literature&33,772&Full contents of articles&Single\\
S2ORC~\cite{bishop2022gencomparesum}&Biomedical literature&63,709&Full contents of articles&Single\\
    CORD-19~\cite{wang2020cord}&Biomedical literature&-&Full contents of articles&Single\\
    PubMedCite~\cite{luo2023citationsum}&Biomedical literature&192,744&Full contents of articles, citation graph&Single\\
CDSR~\cite{guo2021automated}&Biomedical literature&7,805&Abstracts of articles&Single\\
PLOS~\cite{luo2022readability}&Biomedical literature&28,124&Full contents of articles&Single\\
    RCT~\cite{wallace2021generating}&Biomedical literature&4,528&Titles and abstracts of articles&Multiple\\
MSˆ2~\cite{deyoung2021ms2}&Biomedical literature&470,402&Abstracts of articles&Multiple\\
    MIMIC-CXR~\cite{johnson2019mimic}&EHRs&124,577&Full contents of reports&Single\\
    OpenI~\cite{demner2016preparing}&EHRs&3,599&Full contents of reports&Single\\
    HET-MC~\cite{song2020summarizing}&Medical conversation&109,850&Multi-turn conversations &Single\\
    MeQSum~\cite{abacha2019summarization}&Medical question&1,000&Consumer health question&Single\\
    CHQ-Summ~\cite{yadav2022chq}&Medical question&1,507&Full contents of question&Single\\
    \bottomrule
    \end{tabular}
    \caption{Biomedical text summarization datasets.}
    \label{tab:bio}
\end{table*}

\textbf{Biomedical Literature} 
With the exponentially growing of scientific papers, developing automated summarization tools for biomedical articles has long attracted much attention.
Biomedical articles are usually written by domain experts such as researchers and physicians.
Compared with general domain texts such as social media texts or news texts, they are typically less noisy and exhibit a structured organization with standardized sections such as "Introduction," "Methods," "Results," and others.

For single document summarization, PubMed~\cite{cohan2018discourse} is one of the most commonly used datasets, for summarization of long biomedical texts.
It consists of 133K scientific papers collected from the PubMed open access repositories\footnote{\url{https://www.ncbi.nlm.nih.gov/pmc/tools/openftlist/}}.
It has become a popular benchmark dataset for evaluating both general text summarization methods and biomedical text summarization techniques.
It is noticed that Zhong et al~\cite{zhong2020extractive} further adapt the dataset that only uses the introduction of texts as the input.
To identify these two settings on the dataset, we name the original PubMed dataset that uses full contents of documents as the PubMed-Long\footnote{\url{https://github.com/armancohan/long-summarization}}, and the new dataset that is adapted by Zhong et al~\cite{zhong2020extractive} as the PubMed-Short.
Following it, SumPubMed~\cite{gupta2021sumpubmed} proposed recently, includes 33,772 documents from Bio Med Central (BMC) of PubMed archive\footnote{\url{https://github.com/vgupta123/sumpubmed}}.
Bishop et al \cite{bishop2022gencomparesum,xie2022pre} extract the subset from the large scientific corpus S2ORC~\cite{lo2020s2orc} and build the dataset S2ORC\footnote{\url{https://github.com/jbshp/GenCompareSum}} which includes 63,709 articles from the biological and biomedical domain.
Most recently, COVID-19 Open Research Dataset (CORD-19\footnote{\url{https://github.com/allenai/cord19}})~\cite{wang2020cord} has attracted much attention, for which developing summarization systems would facilitate relevant research and help against the COVID-19 pandemic.
CORD-19 contains millions of papers related to COVID-19, SARS-CoV-2, and other coronaviruses.
Luo et al~\cite{luo2023citationsum} propose the PubMedCite\footnote{\url{https://github.com/zhehengluoK/PubMedCite-Builder}} dataset with the 192K biomedical scientific papers and a citation graph preserving 917K citation relationships between them, for supporting the biomedical scientific paper summarization with the citation graph.
Moreover, Guo et al~\cite{guo2021automated} collected the CDSR\footnote{\url{https://github.com/qiuweipku/Plain\_language\_summarization}} dataset to support the task of lay language summarization of biomedical scientific reviews, which is a special kind of single document summarization that aims to generate plain language abstract for lay people based on professional abstracts from expertise.
It contains 7,805 abstract pairs of biomedical scientific reviews, in which professional abstracts of systematic reviews are deemed as inputs and their corresponding plain language abstracts as target summaries.
Following it, Luo et al~\cite{luo2022readability} present the PLOS corpus\footnote{\url{http://www.nactem.ac.uk/readability/}} with 28,124 biomedical papers along with their technical summaries and plain summaries.

For multi-document summarization in the biomedical domain,
Wallace et al \cite{wallace2021generating} build the RCT\footnote{\url{https://github.com/bwallace/RCT-summarization-data}} summarization dataset with 4,528 data samples searched from PubMed\footnote{\url{https://pubmed.ncbi.nlm.nih.gov}}.
The input of each data sample includes titles and abstracts of related papers describing randomized controlled trials (RCTs), while the conclusion section of the systematic review from Cochrane\footnote{\url{https://www.cochranelibrary.com/}} is treated as the target summary.
Similarly, Deyoung et al \cite{deyoung2021ms2} developed the MSˆ2\footnote{\url{https://github.com/allenai/ms2/}} for multi-document summarization of medical studies.
It collected 470K papers from Semantic Scholar and 20K reviews that summarized these papers.

\textbf{Electronic Health Records} 
Electronic health records have been widely adopted by hospitals to store and manage medical information of patients, such as diagnostic codes, medications, laboratory results, clinical notes et al.
They are also written by professionals with professional language and specific structure.
Different from scientific papers that are generally free to access,
EHRs may have restrictions on public access due to privacy issues.
Several publicly available datasets have been released to support the automated summarization of radiology reports.
The task aims to automatically generate the impressions which highlight key observations of the radiology reports.
Demner et al \cite{demner2016preparing} collected the OpenI\footnote{\url{https://openi.nlm.nih.gov/faq\#collection}} datasets containing 3,996 chest x-ray reports from hospitals within the Indiana Network.
Compared with OpenI, MIMIC-CXR~\cite{johnson2019mimic} is a larger publicly available dataset including 107,372 radiology reports from Beth Israel Deaconess Medical Center Emergency Department between 2011–2016\footnote{\url{https://physionet.org/content/mimic-cxr/2.0.0/}}.

\textbf{Medical Conversations} 
Medical conversations between patients and doctors from online healthcare systems have become an important source of medical information, with the increasing usage of telemedicine.
The automated summarizing of key medical information during long medical conversations can save much time for doctors and improve healthcare efficiency.
Medical conversations usually involve multi-turn interactions between two parties.
The patients focus on asking questions and solutions to their health problems and describing their symptoms, while doctors would ask for detailed symptoms of patients and provide diagnostic suggestions.
Similar to EHRs, accessing medical conversations at telemedicine platforms may have restrictions due to privacy concerns.
Moreover, it is time-consuming and expensive to build the training data, since it requires professionals to write target summaries manually.
Up to now, although several advanced methods with PLMs for medical conversation summarization have been proposed~\cite{enarvi2020generating,song2020summarizing,chintagunta2021medically,zhang2021leveraging,yim2021towards}, publicly available datasets are limited.
Song et al \cite{song2020summarizing} proposed the Chinese medical conversation summarization dataset\footnote{\url{https://github.com/cuhksz-nlp/HET-MC}} with 109,850 conversations from the online health platform\footnote{\url{https://www.chunyuyisheng.com/}}.

\textbf{Medical Questions} 
The consumer health questions produced by healthcare consumers on the web such as health forums, are another important data source of clinical information.
To find trustworthy answers to their health questions, healthcare consumers can query the web with long natural language questions with peripheral details.
The peripheral information is useless to find high-quality answers to health questions.
Therefore, summarizing consumer health questions into concise text with salient information is quite useful for improving the efficiency of medical question answering.
Abacha et al \cite{abacha2019summarization} build the MeQSum\footnote{\url{https://github.com/abachaa/MeQSum}} corpus with 1,000 consumer health questions as inputs and their manual summaries from three medical experts.
Yadav et al \cite{yadav2022chq} introduced another dataset CHQ-Summ\footnote{\url{https://github.com/shwetanlp/Yahoo-CHQ-Summ}} most recently, which includes 1,507 consumer health questions and their summaries annotated by experts.
Different from other texts such as biomedical papers with thousands of words, consumer health question-summary pairs are short texts.
For example, in CHQ-Summ, the average length of questions and their summaries are 200 words and 15 words respectively.

\section{Methods}
\label{sec:method}
Given biomedical text datasets,
many methods have explored how to make use of PLMs and LLMs for the biomedical summarization task.
Different from previous reviews~\cite{afantenos2005summarization,mishra2014text,pivovarov2015automated,wang2021systematic} for traditional and deep learning methods that classify methods according to their inputs (multiple document/single document) and outputs (extractive/abstractive), we categorize related methods based on how PLMs and LLMs are leveraged in recent research to improve biomedical text summarization.

\subsection{PLMs for Biomedical Text Summarization}
For PLMs-based methods, we first categorize them into three major categories: feature-based, fine-tuning-based, and domain-adaption-with-fine-tuning-based, according to the ways that they introduce PLMs into biomedical text summarization.
We then categorize methods according to the structure of PLMs and the dataset they adopt.

\begin{figure*}
    \centering
    \subfloat[The framework of the feature-based method.]{
   \begin{minipage}{.35\textwidth}
  \centering
  \includegraphics[width=.99\linewidth]{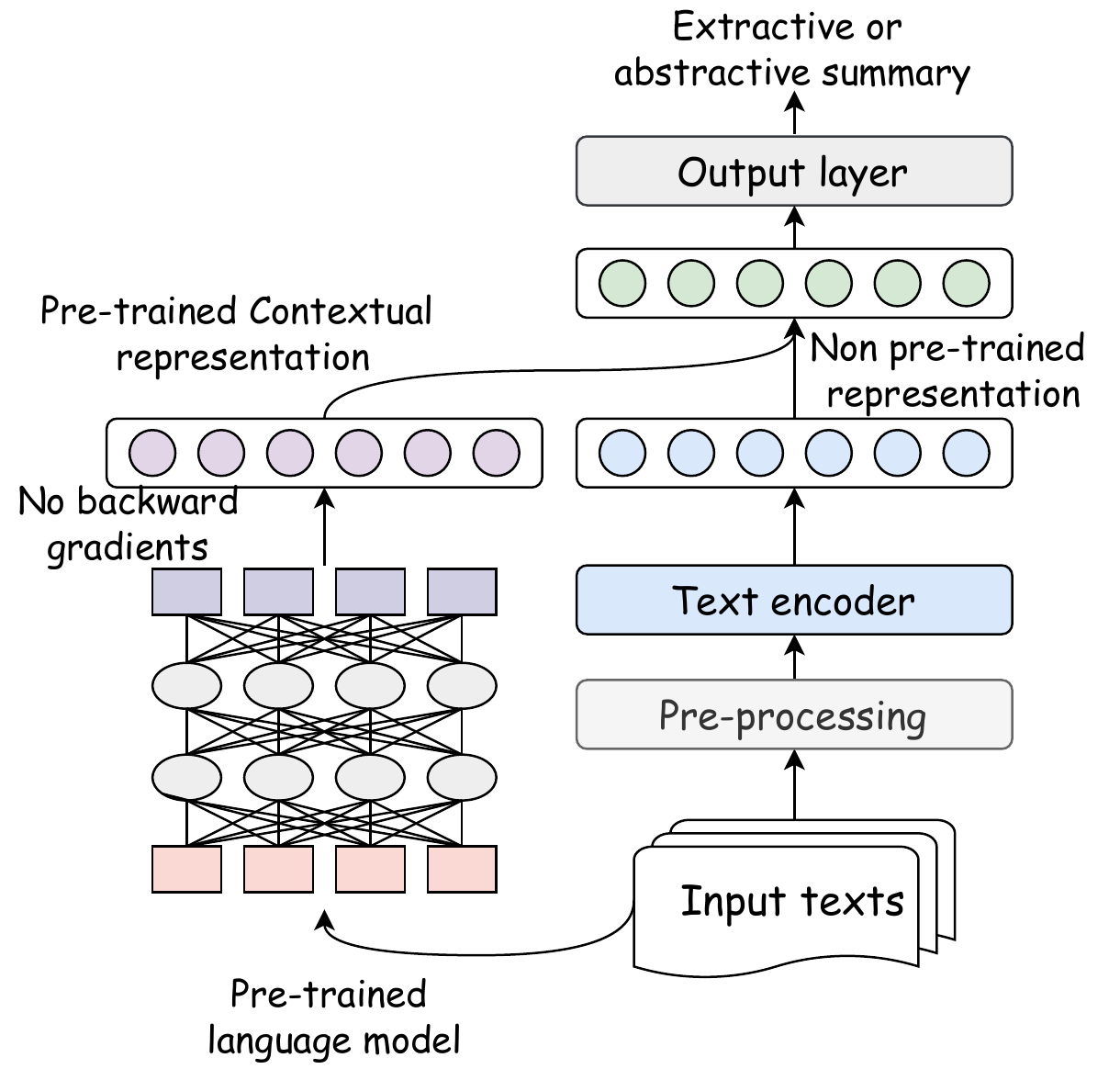}
\end{minipage}%
}
\subfloat[The framework of the fine-tuning-based method.]{
\begin{minipage}{.20\textwidth}
  \centering
  \includegraphics[width=.99\linewidth]{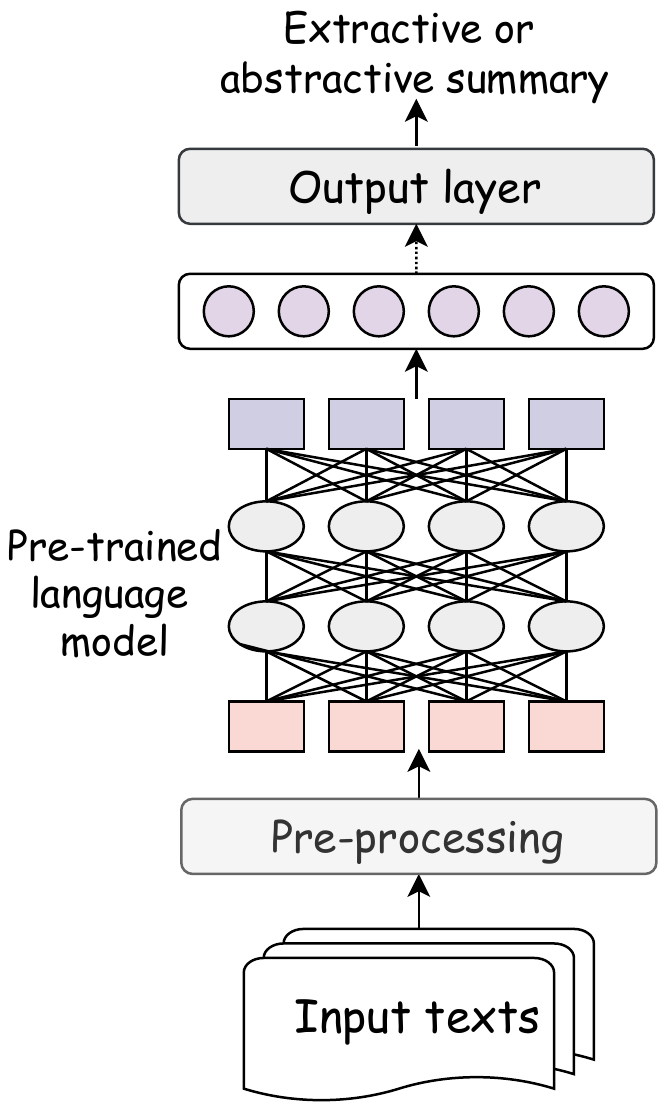}
\end{minipage}
}
\subfloat[The framework of domain-adaption + fine-tuning based method.]{
\begin{minipage}{.40\textwidth}
  \centering
  \includegraphics[width=.99\linewidth]{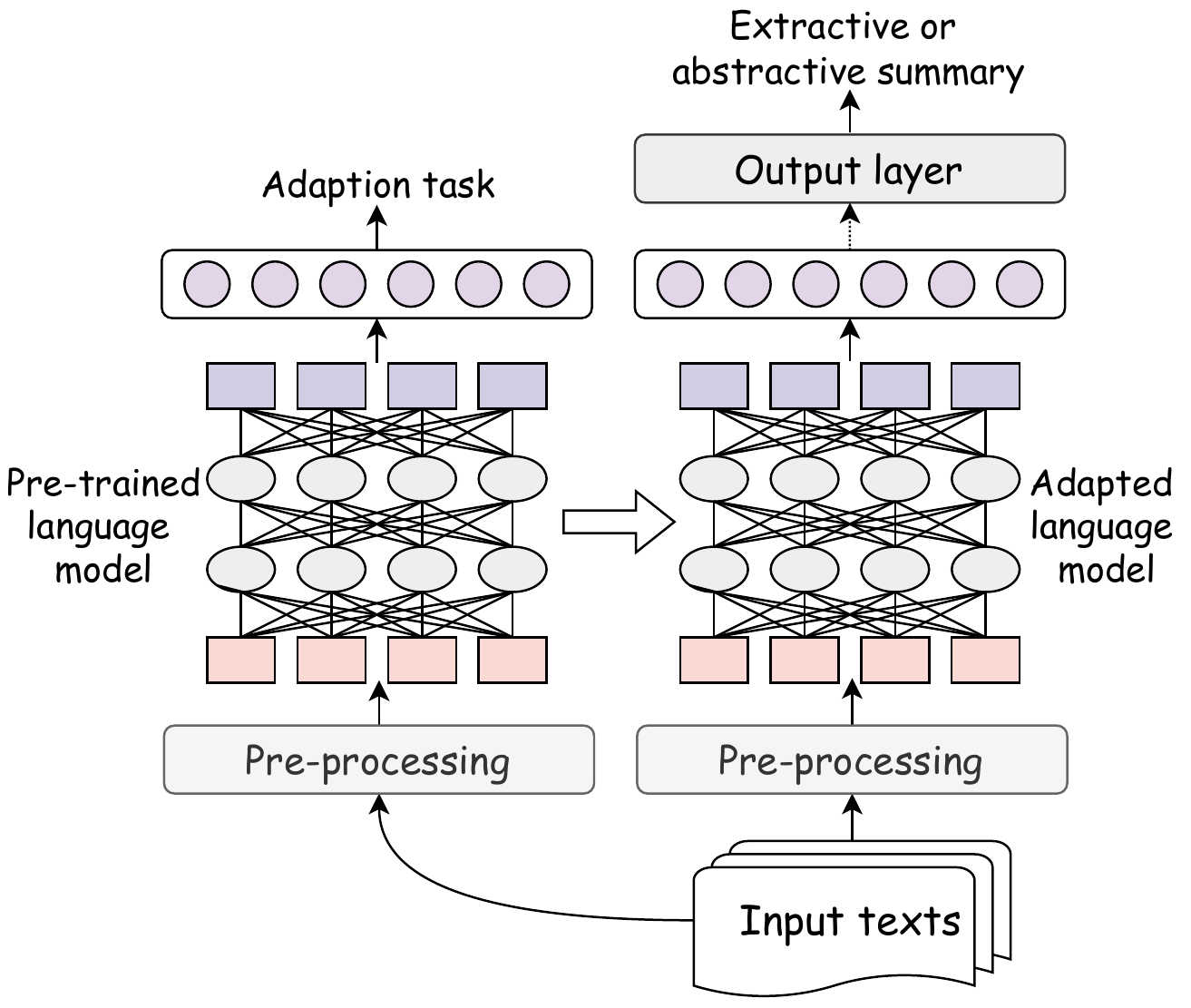}
\end{minipage}
}
\caption{Comparison of different strategies on using PLMs for BTS.}
\label{fig:com}
\end{figure*}

As shown in Figure \ref{fig:com},
the feature-based methods independently utilize contextual representations from PLMs without refining the pre-trained parameters of PLMs.
The representations from PLMs are generally concatenated with the representations from the encoder to generate the output.
Although they are time-saving for fixing the parameters of PLMs, their performance is limited since they do not consider task-specific supervised information.
The fine-tuning-based methods employ PLMs as text encoders, where the generated representations are directly fed into the output layer. The parameters of the PLMs, along with task-specific parameters, are fine-tuned based on the task loss. These methods often require higher computing resources and are time-consuming, but they offer improved and promising performance compared to feature-based methods.
The fine-tuning-with-domain-adaption-based methods first conduct the domain-adaption for PLMs via continually training PLMs with designed tasks on the target data.
The adapted PLMs, along with task-specific layers, are then fine-tuned. 
This domain adaptation process enables the PLMs to better capture domain-specific knowledge, resulting in improved performance compared to methods directly fine-tuning PLMs.
We will review and discuss these methods in more detail, as shown in Table \ref{tab:com}.
\begin{table*}[htb]
    \centering
    \scriptsize
    \begin{tabular}{c|ccccccc}
    \toprule  \bf{Paper}&\bf{Strategy}&\bf{Model}&\bf{Category}&\bf{Input}&\bf{Output}&\bf{Training}&\bf{Data}\\
    \midrule   
    Moradi et al\cite{moradi2020deep}&feature-base&encoder&literature&single&extractive&unsupervised&-\\
    Moradi et al\cite{moradi2020summarization}&feature-base&encoder&literature&single&extractive&unsupervised&-\\
Gharebagh et al\cite{gharebagh2020attend}&feature-base&encoder&EHRs&single&abstractive&supervised&MIMIC-CXR, OpenI\\
RadBERT\cite{yan2022radbert}&feature-base&encoder&EHRs&single&extractive&unsupervised&-\\
GenCompareSum~\cite{bishop2022gencomparesum}&feature-base&encoder-decoder&literature&single&extractive&unsupervised&PubMed, CORD-19, S2ORC\\
Su et al \cite{su2020caire}&fine-tuning,feature-base&encoder, encoder-decoder&literature&mutiple&hybrid&un+supervised&CORD-19\\
    ContinualBERT~\cite{park2020continual}&fine-tuning& encoder&literature&single&extractive&supervised&CORD-19\\
    BioBERTSum~\cite{du2020biomedical}&fine-tuning&encoder&literature&single&extractive&supervised&-\\
    Cai et al~\cite{cai2022covidsum}&fine-tuning&encoder&literature&single&abstractive&supervised&CORD-19, PubMed\\
     Kanwal et al\cite{kanwal2022attention}&fine-tuning&encoder&EHRs&single&extractive&unsupervised&MIMIC-III\\
      Hu et al~\cite{jingpeng2022graph}&fine-tuning&encoder&EHRs&single&abstractive&supervised&MIMIC-CXR,OpenI\\
       HET~\cite{song2020summarizing}&fine-tuning&encoder&conversation&single&extractive&supervised&HET-MC\\
    Esteva et al \cite{esteva2021covid}&fine-tuning&decoder&literature&multiple&abstractive&supervised&CORD-19\\
     Deyoung et al \cite{deyoung2021ms2}&fine-tuning&encoder-decoder&literature&multiple&abstractive&supervised&MSˆ2\\
     Luo et al \cite{luo2022readability}&fine-tuning&encoder-decoder&literature&single&abstractive&supervised&PLOS\\
    Zhu et al \cite{zhu2021paht_nlp}&fine-tuning&encoder-decoder&EHRs&single&abstractive&supervised&MIMIC-CXR,OpenI\\
    Kondadadi et al \cite{kondadadi2021optum}&fine-tuning&encoder-decoder&EHRs&single&abstractive&supervised&MIMIC-CXR,OpenI\\
     FactReranker~\cite{xie2023factreranker}&fine-tuning&encoder-decoder&EHRs&single&abstractive&supervised&MIMIC-CXR,OpenI\\
     Xu et al \cite{xu2021chichealth}&fine-tuning&encoder-decoder&EHRs,question&single&abstractive&supervised&MIMIC-CXR,OpenI,MeQSum\\
     He et al \cite{he2021damo_nlp}&fine-tuning&encoder-decoder&EHRs,question&single&abstractive&supervised&MIMIC-CXR,OpenI,MeQSum\\
      Yadav et al~\cite{yadav2021reinforcement}&fine-tuning&encoder-decoder&question&single&abstractive&supervised&MeQSum\\
CLUSTER2SENT~\cite{krishna2021generating}&fine-tuning&encoder-decoder&conversation&single&abstractive&supervised&-\\
    Zhang et al~\cite{zhang2021leveraging}&fine-tuning&encoder-decoder&conversation&single&abstractive&supervised&-\\
    Navarro et al~\cite{navarro2022few}&fine-tuning&encoder-decoder&conversation&single&abstractive&supervised&-\\
    BioBART~\cite{yuan2022biobart}&fine-tuning&encoder-decoder&conversation&single&abstractive&supervised&-\\
     KeBioSum~\cite{xie2022pre}&adaption+fine-tuning&encoder&literature&single&extractive&supervised&PubMed, CORD-19, S2ORC\\
    Yalunin et al~\cite{yalunin2022abstractive}&adaption+fine-tuning&encoder&EHRs&single&abstractive&supervised&-\\
    Mahajan et al~\cite{mahajan2021ibmresearch}&adaption+fine-tuning&encoder&EHRs&single&abstractive&supervised&MIMIC-CXR,OpenI\\
     Yadav et al~\cite{yadav2022question}&adaption+fine-tuning&encoder&question&single&abstractive&supervised&MeQSum\\
      Kieuvongngam et al~\cite{kieuvongngam2020automatic}&adaption+fine-tuning&decoder&literature&single&hybrid&supervised&CORD-19\\
      Guo et al~\cite{guo2021automated}&adaption+fine-tuning&encoder, encoder-decoder&literature&single&hybrid&supervised&CDSR\\
      Wallace et al~\cite{wallace2021generating}&adaption+fine-tuning&encoder-decoder&literature&multiple&abstractive&supervised&RCT\\
    BDKG~\cite{dai2021bdkg}&adaption+fine-tuning&encoder-decoder&EHRs&single&abstractive&supervised&MIMIC-CXR,OpenI\\
    Mrini et al~\cite{mrini2021gradually}&adaption+fine-tuning&encoder-decoder&question&single&abstractive&supervised&MeQSum\\
    \bottomrule
    \end{tabular}
    \caption{Overview of PLMs-based Methods. Besides how they use PLMs and what PLMs they used, we also list other features of these methods, according to types of output summary (extractive, abstractive, hybrid), numbers of input documents (single, multiple), and types of input documents (biomedical literature summarization, radiology report summarization, medical dialogue summarization, and medical question summarization). "-'' means datasets that are not released.}
    \label{tab:com}
\end{table*}

\subsubsection{\textbf{Feature-based Methods}}
Feature-based methods leverage the contextualized representations of pre-trained language models (PLMs) as additional features in the summarization process. These contextualized representations provide extra semantic information that can't be captured by text encoders based on non pre-trained neural networks.

\textbf{Bi-directional encoder language models}
For biomedical literature texts, Moradi et al \cite{moradi2020deep} propose the unsupervised extractive summarizer based on hierarchical clustering and PLMs.
They conduct sentence clustering based on sentence representations from PLMs and then select top sentences from clusters with the ranking method.
They investigate different versions of BERT and BioBERT to yield sentence representations.
The proposed methods show better performance than traditional unsupervised methods such as TextLexAn\footnote{\url{http://texlexan.sourceforge.net}}.
They find that all versions of BioBERT (BioBERT-pmc, BioBERT-pubmed, BioBERT-pubmed+pmc) outperform the BERT-base, but underperform the BERT-large.
Moradi et al \cite{moradi2020summarization} propose the graph ranking based method for biomedical text summarization.
They use the contextualized embeddings of BioBERT to represent sentences and build graphs for texts.
The important sentences are identified with the graph ranking
algorithm from text graphs.
The model based on BioBERT-pubmed+pmc achieves better performance than models based on the other two versions of BioBERT: BioBERT-pmc and BioBERT-pubmed.
Su et al \cite{su2020caire} propose the query-focused multi-document summarizer with ALBERT~\cite{lan2019albert} for COVID-19 articles.
They use ALBERT~\cite{lan2019albert} to generate sentence representations and calculate the cosine similarity between sentences and queries to select important sentences.

As for clinical notes, Gharebagh et al \cite{gharebagh2020attend} develop supervised method for abstractive summarization of clinical notes.
They propose to incorporate the contextual embeddings from BERT as the input embeddings of the Bi-LSTM-based encoder. 
It shows better performance than classical sequence-to-sequence-based methods based on recurrent neural networks.
Yan et al \cite{yan2022radbert} propose the radiology-specialized language model RadBERT that is pre-trained on millions of radiology reports.
On the unsupervised extractive summarization of radiology reports, it achieves better performance than other language models including BERT, BioBERT, ClinicalBERT, BlueBERT, and BioMed-RoBERTa~\cite{gururangan2020don}.
Among all variants: RadBERT-BERT-base, RadBERT-RoBERTa, RadBERT-ClinicalBERT et al, the RadBERT-BioMed-RoBERTa achieves the best performance.

\textbf{Encoder-decoder language models}
Bishop et al \cite{bishop2022gencomparesum} present unsupervised extractive summarization method GenCompareSum for biomedical literature.
GenCompareSum uses the T5 generative model to generate key snippets for text sections and selects important sentences with BERTScore~\cite{zhang2019bertscore} between key snippets and sentences.
It outperforms traditional unsupervised methods such as LexRank~\cite{erkan2004lexrank}, TextRank~\cite{mihalcea2004textrank}, and also the SOTA supervised method BERTSum.

\subsubsection{\textbf{Fine-tuning based Methods}} 
Fine-tuning is the most common way to use PLMs for downstream tasks.
Different from feature-based methods that freeze the parameters of PLMs, fine-tuning-based methods refine all parameters of PLMs along with task-specific parameters.

\textbf{Bi-directional encoder language models}
For biomedical literature, Park et al \cite{park2020continual} present the ContinualBERT model for adaptive extractive summarization of covid-19 related literature.
ContinualBERT trains two BERT models with continual learning in order to process texts online.
It shows better performance than the SOTA extractive method BERTSum~\cite{liu2019text}.
Du et al \cite{du2020biomedical} propose BioBERTSum model for extractive summarization of biomedical literature.
It uses token embedding, sentence embedding, and position embedding to embed input texts, and then yields contextual representations of sentences with BioBERT.
BioBERT and the extra classifier layer are fine-tuned with the cross-entropy loss.
It proves the advantage of using a domain-specific language model for biomedical texts and outperforms the SOTA method BERTSum that uses BERT in the general domain as the encoder.
Cai et al \cite{cai2022covidsum} propose a SciBERT-based abstractive summarization model for COVID-19 scientific papers, that uses the linguistic information of word co-occurrence encoded by graph attention network to enrich the SciBERT encoder.
They find that their method based on the SciBERT encoder outperforms that based on BERT and BioBERT encoder.

As for clinical notes, Kanwal et al \cite{kanwal2022attention} propose the multi-head attention-based method for extractive summarization.
It fine-tunes BERT on the task of predicting and identifying ICD-9 labels on the ICD-9 labeled MIMIC-III discharge notes.
The attention scores of sentences from the last layer of the BERT model are used to select sentences.
Hu et al \cite{jingpeng2022graph} propose the radiology report summarizer that uses BioBERT as text encoder and randomly initialized transformer layers as the decoder.
They use the graph encoder and contrastive learning to incorporate extra knowledge to improve the BioBERT encoder. 

Moreover, Song et al \cite{song2020summarizing} propose the hierarchical encoder-tagger (HET) model for extractive summarization of medical conversation, which includes token-level and utterance-level encoders to encode input long transcripts.
They use the Chinese version of BERT as the token-level encoder.

\textbf{Auto-regressive decoder language models}
There are abstractive methods that focus on fine-tuning auto-regressive decoder language models such as GPT-2 and GPT-3.
For COVID-19-related biomedical articles,
Esteva et al \cite{esteva2021covid} design the parallel encoder-decoder framework for abstractive summarization of multiple COVID-19 articles, that uses BERT as the encoder and GPT-2 as the decoder.
To overcome the problem of limited training data, they propose to use GPT-3 as the few-shot learner and propose the few-shot fine-tuning strategy.

\textbf{Encoder-decoder language models}
There are many studies explore encoder-decoder language models like BART, T5, and PEGASUS~\cite{zhang2020pegasus} for BTS. 
They are pre-trained with a specific objective function tailored for abstractive text summarization, making them well-suited for the summarization task.

For biomedical literature, Deyoung et al \cite{deyoung2021ms2} develop the BART-based method for multi-document summarization of medical studies. 
To encode multi-documents, they investigate two encoders. 
One is using multiple BART encoders to encode multi-documents separately.
Another one is using LongformerEncoderDecoder (LED)~\cite{beltagy2020longformer}, which can encode long inputs up to 16K tokens.
Luo et al~\cite{luo2022readability} fine-tune the LongformerEncoderDecoder (LED) for generating technical and lay summaries for biomedical scientific papers.
Su et al \cite{su2020caire} fine-tune BART for multi-document abstractive summarization for COVID-19 articles based on user queries.

As for EHRs, the MEDIQA 2021 Shared Task~\cite{abacha2021overview} at the BioNLP 2021 workshop introduces the abstractive summarization task for radiology reports and medical question, in which most participating teams propose methods based on encoder-decoder language models.
Among 14 teams that participate in the radiology reports summarization task, 6 of them~\cite{xu2021chichealth,zhu2021paht_nlp,kondadadi2021optum,mahajan2021ibmresearch,he2021damo_nlp} use the encoder-decoder language models such as BART, PEGASUS.
Most of them~\cite{xu2021chichealth,zhu2021paht_nlp} find that fine-tuning PEGASUS achieves the best performance, while Kondadadi et al~\cite{kondadadi2021optum} reports that the best performance is achieved by the BART.
Moreover, they report that adapting PEGASUS on the PubMed corpus can lead to worse performance, which may be due to the gap between biomedical literature and 
medical reports.
Xie et al~\cite{xie2023factreranker} propose a second stage abstractive summarization framework FactReRanker for clinical notes, where the first stage fine-tunes the BART model to generate multiple candidate summaries and the second stage selects the best summary based on their factual consistency score.

For medical conversation summarization, efforts have been proposed to explore the encoder-decoder language models to address challenges such as limited labeled data and long transcripts.
Krishna et al \cite{krishna2021generating} develop CLUSTER2SENT, an extractive-abstractive hybrid method on doctor-patient conversations to generate SOAP notes (long semi-structured clinical summaries). 
T5 model is used in the abstractive module of CLUSTER2SENT.
Zhang et al \cite{zhang2021leveraging} leverage the BART model for automatic summarization of doctor-patient conversations with limited labeled training data.
They propose the multistage fine-tuning strategy to address the input length limitation of BART.
They find that fine-tuning BART can generate summaries of good quality even with limited training data.
Moreover, they also find that the BART-based model significantly outperforms the BigBird~\cite{zaheer2020big} based models that are initialized by RoBERTa-base and PEGASUS-Large.
Navarro et al \cite{navarro2022few} explore fine-tuning BART, T5, PEGASUS with zero-shot and few-shot learning strategies for medical dialogue summarization with small training data.
They find that BART achieves the best performance among these PLMs.
Yuan et al \cite{yuan2022biobart} develop the first encoder-decoder language model BioBART in the biomedical domain, which has shown better performance on medical dialogue summarization than BART.

Moreover, for medical questions, 22 teams participate in the medical question summarization task of the MEDIQA 2021 Shared Task~\cite{abacha2021overview} at the BioNLP 2021 workshop, and all methods are based on fine-tuning encoder-decoder language models.
The best performance is achieved by the ensemble model~\cite{he2021damo_nlp} that re-ranks summary outputs of multiple advanced encoder-decoder language models including BART, T5, and PEGASUS.
Yadav et al \cite{yadav2021reinforcement} present the reinforcement learning based framework for abstractive summarization of medical questions.
They propose two reward functions: the Question-type Identification Reward (QTR) and Question-focus Recognition Reward (QFR), which are optimized via learning  optimal policy defined by BERT.
They show that the encoder-decoder language model ProphetNet~\cite{qi2020prophetnet} with the proposed reward functions has better performance than other PLMs including T5, BART, and PEGASUS.

\subsubsection{\textbf{Domain Adaption with Fine-tuning based Methods}}
There are summarization methods that conduct domain adaption before fine-tuning PLMs, to capture domain and task-specific information.

\textbf{Bi-directional encoder language models}
For biomedical literature, Xie et al \cite{xie2022pre} propose the KeBioSum for the extractive summarization.
It proposes to refine PLMs with the domain adaption tasks of predicting key entities and their types based on the lightweight fine-tuning framework, which aims to incorporate fine-grained medical knowledge into PLMs. 
It proves that although biomedical language models such as BioBERT, and PubMedBERT can capture domain knowledge to some extent, fine-grained medical knowledge is still beneficial to improve language models.
They find that PubMedBERT-based methods outperform methods based on BERT, RoBERTa, and BioBERT.

As for EHRs, Yalunin et al \cite{yalunin2022abstractive} present the abstractive summarizer for patient hospitalization histories, that uses Longformer~\cite{beltagy2020longformer} as the encoder and BERT as the decoder. 
They propose to pre-train BERT and Longformer with the masked language task on the hospitalization history dataset before task specific fine-tuning.
Dai et al~\cite{dai2021bdkg} propose the BDKG method, which achieves the best performance on MEDIQA 2021 Shared Task for radiology reports summarization.
It ensembles results from multiple language models BART, DistillBERT~\cite{sanh2019distilbert}, PEGASUS, and uses other strategies including domain adaption and text normalization.

Moreover, for medical questions, Yadav et al \cite{yadav2022question} investigate to incorporate the knowledge of "question-focus" and "question-type" with PLMs for abstractive summarization of consumer health questions.
To induce PLMs to capture this knowledge, they adapt PLMs with designed Cloze tasks to capture the .

\textbf{Auto-regressive decoder language models}
Kieuvongngam et al \cite{kieuvongngam2020automatic} use the GPT-2~\cite{radford2019language} for abstractive summarization of COVID-19 medical research articles.
They take keywords of articles as inputs and fine-tune GPT-2 on multi-tasks including the language modeling task and the multiple choice prediction task.

\textbf{Encoder-decoder language models}
For biomedical articles, Guo et al \cite{guo2021automated} use the BART for automated lay language summarization of biomedical review articles.
They conduct domain-adaption before fine-tuning, which pre-trains the BART model to reconstruct original PubMed abstracts with disrupted abstracts.
Wallace et al \cite{wallace2021generating} propose the multi-document abstractive summarization models based on BART for randomized controlled trials (RCTs).
They adapt the BART with the domain-specific pre-training strategy of generating summaries from full-text articles before fine-tuning.
They also use the "decoration'' strategy to explicitly inform key trial attributes (the "PICO'' elements) of input articles.

As for medical questions, Mrini et al \cite{mrini2021gradually} present the multi-task learning and data augmentation method on medical question summarization and recognizing question entailment (RQE) for medical question understanding.
They prove that the multi-task learning between question summarization and RQE is able to increase the performance of PLMs including BART and BioBERT. 
\begin{table*}[t]
\scriptsize
    \centering
    \begin{tabular}{c|c|ccc}
    \toprule
    \bf{Dataset}& \bf{Category}&\multicolumn{3}{|c}{\bf{Metrics}}\\
    \midrule
    \quad&\quad&\multicolumn{3}{|c}{\bf{Automatic metrics}}\\
    \midrule
    \quad&\quad&similarity&factuality&others\\
    \midrule
 PubMed~\cite{cohan2018discourse}&literature&ROUGE~\cite{bishop2022gencomparesum,xie2022pre,cai2022covidsum,park2020continual},BERTScore~\cite{xie2022pre}&-&-\\
    SumPubMed~\cite{gupta2021sumpubmed}&literature&ROUGE~\cite{gupta2021sumpubmed}&-&-\\   
    S2ORC~\cite{bishop2022gencomparesum}&literature&ROUGE~\cite{bishop2022gencomparesum,xie2022pre,cai2022covidsum,park2020continual},BERTScore~\cite{xie2022pre}&-&-\\
    CORD-19~\cite{wang2020cord}&literature&ROUGE~\cite{bishop2022gencomparesum,xie2022pre,cai2022covidsum,park2020continual},BERTScore~\cite{xie2022pre}&-&-\\
    CDSR~\cite{guo2021automated}&literature&ROUGE~\cite{guo2021automated}&-&readability~\cite{guo2021automated}\\
    PLOS~\cite{luo2022readability}&literature&ROUGE~\cite{luo2022readability}&-&readability~\cite{luo2022readability}\\
    PubMedCite~\cite{luo2023citationsum}&literature&ROUGE~\cite{luo2023citationsum}&-&-\\
    RCT~\cite{wallace2021generating}&literature&ROUGE~\cite{wallace2021generating}&-&-\\
    MSˆ2~\cite{deyoung2021ms2}&literature&ROUGE~\cite{deyoung2021ms2}&$\Delta$EI~\cite{deyoung2021ms2}&-\\
    MIMIC-CXR~\cite{johnson2019mimic}&EHRs&ROUGE~\cite{gharebagh2020attend,jingpeng2022graph,zhu2021paht_nlp,kondadadi2021optum,xu2021chichealth,he2021damo_nlp,mahajan2021ibmresearch,dai2021bdkg},BERTScore~\cite{xu2021chichealth,he2021damo_nlp,mahajan2021ibmresearch,dai2021bdkg}&CheXbert~\cite{jingpeng2022graph,xu2021chichealth,he2021damo_nlp,mahajan2021ibmresearch,dai2021bdkg}&-\\
OpenI~\cite{demner2016preparing}&EHRs&ROUGE~\cite{gharebagh2020attend,jingpeng2022graph,zhu2021paht_nlp,kondadadi2021optum,xu2021chichealth,he2021damo_nlp,mahajan2021ibmresearch,dai2021bdkg},BERTScore~\cite{xu2021chichealth,he2021damo_nlp,mahajan2021ibmresearch,dai2021bdkg}&CheXbert~\cite{jingpeng2022graph,xu2021chichealth,he2021damo_nlp,mahajan2021ibmresearch,dai2021bdkg}&-\\
    HET-MC~\cite{song2020summarizing}&conversation&ROUGE~\cite{song2020summarizing}&-&-\\    MeQSum~\cite{abacha2019summarization}&question&ROUGE~\cite{yadav2021reinforcement,yadav2022question,mrini2021gradually}&-&-\\
    CHQ-Summ~\cite{yadav2022chq}&question&ROUGE,BERTScore~\cite{yadav2022chq}&-&-\\
    \bottomrule
    \end{tabular}
    \caption{The usage of automatic evaluation metrics on different biomedical datasets and methods.}
    \label{tab:eval}
\end{table*}
\subsection{LLMs for Biomedical Text Summarization}
Inspired by the impressive capabilities of large language models (LLMs) in various natural language processing (NLP) tasks, there is a growing interest in exploring LLMs for biomedical text summarization. These approaches can be broadly classified into three categories: data augmentation-based methods, zero-shot-based methods, and domain adaption-based methods, based on how they leverage LLMs.
Data augmentation-based methods utilize LLMs as generators to produce high-quality training data. By leveraging the generation capabilities of LLMs, these methods augment the supervised biomedical text summarizers with additional high-quality data, thereby enhancing their performance.
Zero-shot (and few-shot) methods directly prompt LLMs for zero-shot or few-shot biomedical text summarization. Through natural language instructions, these methods enable LLMs to generate summaries without explicit training on the target task.
Domain adaption-based methods focus on continual pre-training or instruction fine-tuning of publicly available LLMs using biomedical datasets or task-specific datasets. 
These approaches aim to enable LLMs to capture medical knowledge and task-specific information, enhancing their effectiveness in biomedical text summarization tasks.

\textbf{Data Augmentation based Methods.}
For medical dialogue summarization, Chintagunta et al~\cite{chintagunta2021medically} integrate medical knowledge and GPT-3.
They consider GPT-3 as the summary generator and choose the best summary that captures the most medical concepts.
They show that GPT-3 can be a promising backbone method for generating high-quality training data that can be incorporated with the training data based on human annotation.

\textbf{Zero-shot based Methods.}
Shaib et al~\cite{shaib2023summarizing} show GPT-3 performs well in summarizing a single medical article but struggles in summarizing multiple biomedical articles in the zero-shot setting.
For radiology reports summarization, Ma et al~\cite{ma2023impressiongpt} propose the ImpressionGPT framework, instructing ChatGPT with dynamically updated prompts, which outperforms supervised methods without additional fine-tuning of LLMs.

\textbf{Domain adaption based Methods.}
For radiology reports summarization, Karn et al~\cite{karn2023shs} pre-train an instruction-tuned LLM Bloomz~\cite{muennighoff2022crosslingual} with radiology reports, which outperforms other supervised methods.
Veen et al \cite{van2023radadapt} investigate adapting LLMs to the task of radiology reports summarization (RRS), by pre-training them with clinical texts and fine-tuning them with examples of RRS. They found that CLIN-T5~\cite{lehman2023clinical} achieves the best performance among other LLMs including FLAN-T5~\cite{chung2022scaling}, SCIFIVE~\cite{phan2021scifive}, and CLIN-T5-SCI~\cite{lehman2023clinical}.

\section{Evaluations}
\label{sec:eval}
Assessing the quality of summaries for biomedical texts poses unique challenges compared to general documents, as biomedical texts tend to be more technical and complex in terms of length and structure. To evaluate the effectiveness of summarization methods, we categorize evaluation metrics into two groups: automatic metrics and human evaluation, based on the level of human involvement.
Automatic metrics in biomedical text summarization can provide performance measurements without requiring human effort. These metrics typically rely on lexical and syntactical information to assess the quality of generated summaries compared to gold (reference) summaries.
In contrast, human evaluation can capture and model semantic information, which is often challenging to quantify. 
However, conducting human evaluation in the biomedical domain is more time-consuming and financially demanding compared to evaluations in the general domain due to the specific domain expertise required.

We further divide existing metrics into three different classes, consisting of similarity, factuality, and others, as shown in Table \ref{tab:eval}.
Similarity metrics focus on the relevance of generated summaries with gold summaries, which are generally based on the overlapping of tokens, phrases, and sentences.
Factuality metrics verify the factual agreement of generated summaries with original documents, which is a critical measurement for the real application of automatic systems, especially in the biomedical domain.
Moreover, there are other metrics such as 1) readability: how easily can human readers understand generated summaries, 2) fluency: how fluent and coherent generated summaries are, and 3) grammaticality: how grammatically correct generated summaries are.

\subsection{Automatic metrics}
\textbf{Similarity}
Similar to the general domain, ROUGE~\cite{lin2004rouge} is the most widely used metric for evaluating the relevance between generated summaries and gold summaries in the biomedical domain,
including (1) ROUGE-1: calculating unigram overlap between generated summaries of summarizers and gold summaries; (2) ROUGE-2: calculating bigram overlap between generated summaries of summarizers and gold summaries; and (3) ROUGE-L: calculating the longest common subsequences between generated summaries of summarizers and gold summaries.
However, ROUGE metrics are limited to relying on shallow lexical overlaps without considering the paraphrasing and terminology variations when measuring similarity.
To address this issue, BERTScore~\cite{zhang2019bertscore} metric has been proposed, that calculates the similarity between two sentences as the sum of cosine similarities between the contextual embeddings of their tokens from pre-trained language models.

\textbf{Factuality}
Compared with extractive methods, it is reported~\cite{zhang2018learning} that abstractive methods struggle to generate factual correct summaries.
It has been a growing awareness that metrics such as ROUGE and BERTScore can not reflect the factual correctness of generated summaries~\cite{zhang2020optimizing}.
For factuality evaluation, Deyoung et al~\cite{deyoung2021ms2} propose the $\Delta$EI metric to calculate the factual agreement of generated summaries and input medical studies.
They propose to calculate the Jensen-Shannon Distance (JSD) between distributions of generated summaries and input medical studies in three directions (increase, decrease, no change) of reported directionality.
However, the metric can only be used in the MSˆ2 dataset.
Zhang et al~\cite{zhang2020optimizing} propose the factual F1 score to evaluate the factual correctness of generated summaries of radiology reports.
They propose to use the CheXbert labeler~\cite{smit2020combining} to yield the binary presence values of disease variables of generated summaries and references and then calculate the overlap of yielded binary presence values between them.

\textbf{Others}
Guo et al~\cite{guo2021automated} propose to apply the readability evaluation which verifies if the generated summaries are understandable for laymen.
It utilizes three different metrics including Flesch-Kincaid grade level~\cite{kincaid1975derivation}, Gunning fog index~\cite{gunning1952technique}, and Coleman-Liau index~\cite{coleman1975computer}.

\subsection{Human evaluation}
While manual evaluation of summarization methods is time-consuming and costly, it provides valuable insights that cannot be captured by automated evaluation metrics. Human evaluation when conducted by domain experts, can assess various aspects of summaries, including fluency, coherence, factuality, grammaticality, and more.
Generally, the human evaluation would recruit human evaluators which are able to read and write in English and participate in medical training and biology courses.
Evaluators are required to score the generated summaries with designed questions that focus on one of the aforementioned aspects.
Moramarco et al \cite{moramarco2022human} study the correlation between human evaluation and 18 automatic evaluation metrics including text overlap metrics such as ROUGE, CHRF~\cite{popovic2015chrf}, embedding metrics BertScore, and factual F1 score et al, on generated clinical consultation notes.
They find that simple character-based metrics such as character-based Levenshtein distance can be more effective than other complex metrics such as BERTScore, and the choice of human references can largely influence the performance of automatic evaluation metrics.

\textbf{Similarity}
For lay summarization of biomedical literature, Guo et al~\cite{guo2021automated} also propose to conduct human evaluation to further assess the quality of generated summaries on the CDSR dataset.
Different from other research, it requires human evaluators to not participate in medical training and biology courses since its method is designed for laymen.
It proposes the meaning preservation metric for human evaluators, which requires them to answer whether the generated summaries cover the key information of source documents, on the 1-5 Likert scale (1 means very poor, 5 means very good).
For multi-document summarization dataset RCT, Wallace et al~\cite{wallace2021generating} ask evaluators who are medical doctors to score the relevance of the generated summaries to the given topic from mostly off-topic to strongly on-topic.

For medical question summarization, Yadav et al~\cite{yadav2021reinforcement} request two experts in medical informatics to measure the semantics preserved in the generated summaries, i.e, whether the question intent was mentioned in the generated summary.

\textbf{Factual consistency}
For biomedical literature, Guo et al~\cite{guo2021automated} ask evaluators to judge the factuality of the generated summaries on the CDSR dataset following the 1-5 Likert scale.
Wallace et al~\cite{wallace2021generating} request evaluators to answer two questions about the factuality of generated summaries for RCT which concern the factuality degree of the generated summaries compared with gold summaries.
Otmakhova et al~\cite{otmakhova2022patient} define three different metrics for the factuality of the generated summaries on the MSˆ2 dataset, including (1) PICO correctness: the generated summary should contain the same patient population, intervention, and outcome (which are the entity types defined by PICO) as the gold summary; (2) direction correctness: the generated summary should have the same direction referring to the intervention's effect to the outcome as the gold summary, which can be classified as a positive effect, negative effect, and no effect; (3) modality: the confidence of the generated summary about the claim should be the same as the gold summary, which can be defined as a strong claim, moderate claim, weak claim, no evidence, and no claim.

As for medical questions,
Yadav et al~\cite{yadav2021reinforcement} ask experts to verify if all key entities appear in the generated summaries on RCT to assess their factual consistency.

\textbf{Other} 
There are several methods that propose new human evaluation metrics to assess the performance of biomedical summarizers.
Moramarco et al \cite{moramarco2021towards} propose an objective human evaluation based on counting medical facts for generated summaries of medical reports.
Guo et al~\cite{guo2021automated} conduct human evaluation based on questions for grammatical correctness and readability respectively for CDSR dataset.
Wallace et al~\cite{wallace2021generating} ask evaluators to evaluate the fluency of the generated summaries on the RCT dataset.
Otmakhova et al \cite{otmakhova2022patient} consider the fluency of the generated summaries and propose three metrics to evaluate it, including grammatical correctness, lexical correctness, and absence of repetition.
\begin{table*}[tp]
\small
    \centering
    \begin{tabular}{c|ccccc}
    \toprule
    \bf{Methods}& \bf{Strategy}&\bf{Model}&\bf{ROUGE-1}&\bf{ROUGE-2}& \bf{ROUGE-L}\\
    \midrule
    TextRank~\cite{mihalcea2004textrank}&-&-&34.53&12.98&30.99\\
    BERTSum~\cite{liu2019text}&fine-tuning&BERT&34.00&13.42&30.69\\
    PubMedBERTSum~\cite{liu2019text}&fine-tuning&PubMedBERT&34.98&14.22&31.37\\
   KeBioSum~\cite{xie2022pre}&domain adaption+fine-tuning&PubMedBERT&36.39&16.27&33.28\\
    \bottomrule
    \end{tabular}
    \caption{ROUGE F1 score of generated summaries by the SOTA extractive methods on the PubMed-long dataset, that extract 3 sentences to formulate the final summary.}
     \label{tab:pubmed-long}
    \centering
    \begin{tabular}{c|ccccc}
    \toprule
    \bf{Methods}& \bf{Strategy} &\bf{Model}& \bf{ROUGE-1}&\bf{ROUGE-2}& \bf{ROUGE-L}\\
    \midrule
    TextRank~\cite{mihalcea2004textrank}&-&-&38.15&12.99&34.77\\
    BERTSum~\cite{liu2019text}&fine-tuning&BERT&41.09&15.51&36.85\\
   MatchSum~\cite{zhong2020extractive}&fine-tuning&RoBERTa&41.21&14.91&36.75\\
   KeBioSum~\cite{xie2022pre}&domain adaption+fine-tuning&PubMedBERT&43.98&18.27&39.93\\
    \bottomrule
    \end{tabular}
    \caption{ROUGE F1 score of generated summaries by the SOTA extractive methods on the PubMed-short dataset, that extract 6 sentences to formulate the final summary.}
    \label{tab:pubmed}
    \begin{tabular}{c|cc|ccc|ccc}
    \toprule
    \multicolumn{3}{c|}{\bf{Data}}&\multicolumn{3}{c|}{\bf{CORD-19}}&\multicolumn{3}{c}{\bf{S2ORC}}\\
    \midrule
    \bf{Methods}& \bf{Strategy} &\bf{Model}& \bf{ROUGE-1}&\bf{ROUGE-2}& \bf{ROUGE-L}&\bf{ROUGE-1}&\bf{ROUGE-2}& \bf{ROUGE-L}\\
    \midrule
    TextRank~\cite{mihalcea2004textrank}&-&-&32.99&10.39&24.471&36.58&13.23&33.10\\    SumBasic~\cite{nenkova2005impact}&-&-&33.88&8.24&30.86&36.63&10.43&33.68\\
    BERTSum~\cite{liu2019text}&fine-tuning&BERT&38.95&12.17&35.48&43.56&17.85&40.40\\
    GenCompareSum~\cite{bishop2022gencomparesum}&feature-base&T5&41.02&13.79&37.25&43.39&16.84&39.82\\
    \bottomrule
    \end{tabular}
    \caption{ROUGE F1 score of generated summaries by the SOTA extractive methods on the CORD-19 and S2ORC datasets, which extract 8 and 9 sentences for the CORD-19 and S2ORC datasets correspondingly to formulate the final summary.}
    \label{tab:cord19}
    \footnotesize
    \begin{tabular}{c|ccccccccccc}
    \toprule
    \bf{Methods} & \bf{Strategy}&\bf{Model}& \bf{R-1}&\bf{R-2}& \bf{R-L}& \bf{PICO}&\bf{Direction}& \bf{Modality}&\bf{Grammar}&\bf{Lexical}&\bf{Non-redundancy}\\
    \midrule
   BART~\cite{deyoung2021ms2}&fine-tuning&BART&27.56&9.40&20.80&45\%&77\%&45\%&75\%&69\%&85\%\\
   LED~\cite{deyoung2021ms2}&fine-tuning&Longformer&26.89&8.91&20.32&40\%&75\%&44\%&63\%&73\%&89\%\\
    \bottomrule
    \end{tabular}
    \caption{ROUGE F1 score~\cite{lin2004rouge}, factual correctness, grammatical errors and fluency~\cite{otmakhova2022patient} of generated summaries by SOTA abstractive methods for multiple biomedical document summarization on MSˆ2. ROUGE-1, ROUGE-2 and ROUGE-L are commonly used for evaluating the relevancy between gold summaries and generated summaries. PICO, direction, and modality are used for evaluating the factual correctness of generated summaries.}
    \label{tab:multiple}
    \begin{tabular}{c|cc|cccccccc}
    \toprule
    \multicolumn{3}{c|}{\bf{Metrics}}&\multicolumn{3}{c}{\bf{OpenI}}&\multicolumn{5}{c}{\bf{MIMIC-CXR}}\\
        \midrule
    \bf{Method}& \bf{Strategy}&\bf{Model}&\bf{ROUGE-1}&\bf{ROUGE-2}& \bf{ROUGE-L}&\bf{ROUGE-1}&\bf{ROUGE-2}& \bf{ROUGE-L}& \bf{$F_1$CheXbert} & \bf{RadGraph}\\
    \midrule
    LexRank~\cite{erkan2004lexrank}&-&-&14.63                & 4.42  & 14.06 
& 18.11                & 7.47  & 16.87 & -             & -        \\

    TRANSABS~\cite{liu2019text}&-&-&59.66&49.41&59.18&47.16&32.31&45.47&-        & - \\
    Hu et al~\cite{jingpeng2022graph}&fine-tuning&BioBERT& 64.97                & 55.59 & 64.45 
& 51.02                & 35.21 & 46.65 & 70.73         & 45.23    \\
    FactReranker~\cite{xie2023factreranker}&fine-tuning&BART& 66.11&56.31&65.36&55.94&40.63&51.85&76.36&53.17\\ 
    ImpressionGPT~\cite{ma2023impressiongpt}&zero-shot&ChatGPT&66.37&54.93&65.47&54.45&34.50&47.93&-&-\\
    \bottomrule
    \end{tabular}
    \caption{Performance of SOTA abstractive methods for radiology findings summarization on MIMIC-CXR and OpenI.}
    \label{tab:radiology}
    \begin{tabular}{c|ccccc}
    \toprule
    \bf{Method}&\bf{Strategy}&\bf{Model}&\bf{ROUGE-1}&\bf{ROUGE-2}& \bf{ROUGE-L}\\
    \midrule
    Seq2Seq~\cite{sutskever2014sequence}&-&-&25.28&14.39&24.64\\
    BertSum~\cite{liu2019text}&fine-tuning&BERT&26.24&16.20&30.59\\
    PEGASUS~\cite{zhang2020pegasus}&fine-tuning&PEGASUS&39.06&20.18&42.05\\
    Yadav et al \cite{yadav2022question}&domain adaption+fine-tuning&MiniLM&45.20&28.38&48.76\\
    Mrini et al \cite{mrini2021gradually}&domain adaption+fine-tuning&BART&54.5&37.9&50.2\\
    \bottomrule
    \end{tabular}
    \caption{ROUGE F1 score of generated summaries by the SOTA abstractive methods on the MeQSum dataset.}
    \label{tab:meqsum}
\end{table*}
\section{Discussion}
\label{sec:lim}
In this section, we make a further discussion on existing methods and their limitations, and then outlook promising future directions.

\subsection{Comparison}
We first present the performance of existing SOTA PLMs and LLMs based methods on different datasets in Table \ref{tab:pubmed-long}, Table \ref{tab:pubmed}, Table \ref{tab:cord19}, Table \ref{tab:multiple}, Table \ref{tab:radiology}, and Table \ref{tab:meqsum}.

\textbf{How do PLMs and LLMs in biomedical text summarization work?}
For biomedical literature, on the PubMed dataset, as shown in Table \ref{tab:pubmed-long} and Table \ref{tab:pubmed}, PLMs-based methods such as PubMedBERTSum~\cite{liu2019text}, and KeBioSum~\cite{xie2022pre} outperforms TextRank~\cite{mihalcea2004textrank} without PLMs on PubMed-long, and GenCompareSum~\cite{bishop2022gencomparesum}, BERTSum~\cite{liu2019text}, MatchSum~\cite{zhong2020extractive},
and KeBioSum~\cite{xie2022pre} show a significant improvement in both ROUGE metrics compared with TextRank without PLMs on PubMed-short.
It is noticed that the PubMed dataset is also used as a benchmark dataset in the general domain.
Although other advanced methods such as LongT5~\cite{guo2021longt5} have achieved the new SOTA on the dataset, we mainly focus on comparing methods that are designed for the biomedical domain here.
As for CORD-19 and S2ORC, Table \ref{tab:cord19} shows that GenCompareSum~\cite{bishop2022gencomparesum} and BERTSum~\cite{liu2019text} present great performance compared with existing non-PLMs-based methods such as TextRank and SumBasic~\cite{nenkova2005impact}.
Moreover, for multiple document summarization dataset MSˆ2, as shown in Table \ref{tab:multiple}, PLMs-based methods such as BART~\cite{deyoung2021ms2} and LED~\cite{deyoung2021ms2} have shown great performance in both automatical metrics and human evaluation.
We can see that they can generate summaries that have low grammar and lexical error (less than $31\%$), and low redundancy (less than $15\%$).
However, they have problems generating factual correct summaries, for example with PICO correctness no higher than $45\%$.

For radiology reports, Table \ref{tab:radiology} clearly proves the benefits of PLMs via the comparison of PLMs-based methods such as Factreranker~\cite{xie2023factreranker} and non-PLMs-based methods such as TRANSABS~\cite{liu2019text}, and TexRank~\cite{mihalcea2004textrank}.
Factreranker presents the best performance on both MIMIC-CXR and OpenI datasets.
Moreover, the ImpressionGPT which instruct ChatGPT in the zero-shot setting significantly outperforms supervised fine-tuning method such as Hu et al's method~\cite{jingpeng2022graph}, and has comparable performance with Factreranker on both datasets.
This highlights the great potential of LLMs for radiology report summarization in the zero-shot setting.

As shown in Table \ref{tab:meqsum}, PLMs-based methods such as BertSum, PEGASUS~\cite{zhang2020pegasus}, Yadav et al \cite{yadav2022question}, and Mrini et al \cite{mrini2021gradually} both outperform traditional Seq2Seq~\cite{sutskever2014sequence} method on the medical question dataset MeQSum.

Overall, we can find that performance on various datasets has been greatly boosted by advanced methods that make use of PLMs, and existing methods are able to generate fluent summaries, but have limitations on the factual correctness of generated summaries. LLMs have shown great potential for zero-shot BTS without supervised training.

\textbf{What is the optimal way to introduce PLMs and LLMs in biomedical text summarization?}
From Table \ref{tab:pubmed} and Table \ref{tab:meqsum}, we can find that
domain-adaption with fine-tuning based methods such as KeBioSum~\cite{xie2022gretel} and Mrini et al~\cite{mrini2021gradually} present the best performance among three different ways to introduce PLMs.
Fine-tuning based methods generally yield better performance than feature-based methods, except for methods on the CORD-19 and S2ORC datasets, as shown in Table \ref{tab:cord19}.
For multiple biomedical document summarization and EHRs summarization, there are only fine-tuning based methods which are proven to be effective in introducing PLMs.
For radiology reports and medical question summarization, it shows that assembling multiple language models can achieve the best performance~\cite{dai2021bdkg,he2021damo_nlp}. 

\textbf{What is the difference in the choice of PLMs and LLMs?} 
The choice of PLMs and LLMs has a significant influence on the performance of biomedical text summarization.
For biomedical literature, it has been proven that the domain-specific language model BioBERT has better performance than BERT-base, but underperforms BERT-large~\cite{moradi2020deep}.
PubMedBERT further outperforms BioBERT~\cite{xie2022pre}.
As shown in Table \ref{tab:pubmed}, methods using PubMedBERT which is continually pre-trained on the biomedical texts show a better performance compared with general PLMs such as BERT and RoBERTa on both PubMed-long and PubMed-short datasets.
On CORD-19 and S2ORC datasets, the method GenCompareSum with T5 outperforms the method BERTSum with BERT even though T5 applied in their methods would not be fine-tuned.
As for multiple document summarization, LED based on Longformer which is specially designed for long texts presents an inferior performance compared with BART, indicating that it requires more effort to address the input length limitation of PLMs in the biomedical domain rather than directly applying solutions from the general domain.

As for EHRs, medical conversations, and medical questions, most methods use the encoder-decoder language models such as BART, T5, and PEGASUS for abstractive summarization of these datasets.
For EHRs, they find that domain-specific language models such as BioBERT and PubMedBERT, are not effective for radiology reports summarization, since they are pre-trained on biomedical literature~\cite{kondadadi2021optum}.
Language models such as RadBERT which are pre-trained on radiology reports, are better choices~\cite{yan2022radbert}.
For radiology reports summarization, PEGASUS achieves better performance than BART and T5~\cite{xu2021chichealth,zhu2021paht_nlp,dai2021bdkg}.
For medical conversation summarization, it shows that GPT-3 and BART are promising methods with limited training data~\cite{navarro2022few,chintagunta2021medically}.
For medical questions, Mrini et al~\cite{mrini2021gradually} with BART have the best performance compared with other methods based on BERT, PEGASUS, and MiniLM.

\subsection{Limitations}
While methods based on PLMs and LLMs have significantly improved the performance of biomedical text summarization, they still come with a number of limitations.

\textbf{Developing high-quality public datasets}
The development of public datasets for biomedical text summarization is imbalanced.
On one hand, from the perspective of dataset types, compared with a number of public datasets in biomedical literature, there are limited released datasets for electronic health records, medical conversations, and medical questions due to privacy issues, despite the fact that there is an urgent need for developing automated text summarization methods in these texts.
On the other hand, considering the task types, there are only two public datasets for multi-document summarization, while most of the existing datasets focus on single-document summarization.
Moreover, the size of datasets in the biomedical domain such as CHQ-SUmm is generally much smaller than those in the general domain, due to the high cost and time-consuming of human annotation which additionally requires domain-specific knowledge.
The lack of high-quality large-scale public datasets can hinder the development and employment of PLMs whose performance relies on the amount of data.

\textbf{Encoding long biomedical texts} PLMs and LLMs have limitations on the token length of input documents~\cite{beltagy2020longformer} due to the high time and memory consumption of attention computations in Transformer.
Most PLMs and LLMs-based summarization methods directly truncate input documents and only take their first 512 tokens following methods in the general domain.
However, biomedical texts such as biomedical scientific papers, usually have thousands of tokens.
The truncating operation losses useful information in the truncated contents of input documents.
Moreover, it also leads to the loss of long-range dependencies on long biomedical documents.
Although there is a method~\cite{deyoung2021ms2} that investigates using PLMs that support encoding long documents such as Longformer, it shows poor performance when compared with BART on biomedical texts.
Therefore, it still requires more effort to deal with the limitation to encode long biomedical texts based on PLMs and LLMs efficiently.

\textbf{Incorporating domain-specific knowledge} 
Domain-specific knowledge is critical for understanding biomedical texts.
Vocabularies, taxonomies, and ontologies such as UMLS~\cite{bodenreider2004unified} are important sources of biomedical knowledge.
While existing methods with PLMs and LLMs are able to capture lexical knowledge in biomedical texts, they have no knowledge of words or entities that have particular domain-specific importance and their relations. 
Up to now, limited efforts have been proposed to incorporate external domain-specific knowledge for the summarization of biomedical literature and EHRs~\cite{xie2022pre,jingpeng2022graph,xie2023factreranker}.
It is still a limitation for existing methods on other biomedical texts to capture the knowledge of sources such as biomedical concepts, relations between concepts, and lexicographic information et al when leveraging PLMs and LLMs.

\textbf{Controlling factual consistency}
The factual correctness of generated summaries is especially important for the real application of automatic biomedical summarizers.
Despite this, existing methods have traditionally promoted the unrestrained reconstruction of gold standard summaries, without imposing any specific word-level restrictions during the text generation process.
This leads to generated summaries that fabricate facts of original inputs due to freely rephrasing~\cite{kryscinski2020evaluating,zhang2020optimizing}, which may cause medical errors.
It has shown that existing methods based on PLMs such as BART and LLMs such as ChatGPT struggle with maintaining factual correctness in their generated summaries~\cite{deyoung2021ms2,zhang2020optimizing,Xie2023.04.18.23288752}.
This limitation underscores a critical challenge in the current landscape of biomedical text summarization: developing methods with PLMs and LLMs that can effectively control and ensure the factual consistency of the summaries they generate.

\textbf{Interpretability and transparency}
Similar to other deep learning methods, PLMs and LLMs-based methods have the well-known interpretability problem due to their black-box nature.
For end-users, such as clinicians, it becomes challenging to comprehend the reasoning behind a model's selection of specific words or sentences to compose the final summaries. This lack of transparency can be particularly problematic when the model consistently generates errors, making it difficult for users to pinpoint why these inaccuracies occur.
The explanation and transparency of these models, particularly understanding their inner mechanisms and algorithmic functioning, are crucial for building trustworthy applications for users~\cite{tjoa2020survey}. 
However, the challenge of improving the interpretability of these models, specifically when implementing PLMs and LLMs, remains largely unexplored in the current research landscape.

\textbf{Evaluations}
Objective and comprehensive evaluation metrics are important to evaluate summarization methods efficiently and reliably.
Most existing methods only use the ROUGE and BERTScore metrics to evaluate their models automatically similar to methods in the general domain.
However, it has been reported that they are far from reflecting the quality of generated summaries accurately such as factual correctness, and key finding directions.
While some efforts have been made toward developing metrics for evaluating factuality, these typically apply to only one category of biomedical text. Moreover, their effectiveness and correlation with human evaluations remain largely un-investigated, indicating a need for further research in this area.

\subsection{Future Directions}
In this section, we further discuss promising future directions, which we hope can provide guidelines for future research.

\textbf{New large-scale public biomedical datasets}
For biomedical text summarization, the annotation of the dataset is much more expensive and time-consuming than the general domain since the annotators are required to be domain experts.
Moreover, for datasets such as medical conversations and questions, the privacy issue is more critical compared with biomedical literature datasets.
Although there is an urgent need for summarization methods to facilitate information processing, rare attention has been paid to developing high-quality large-scale public datasets for biomedical summarization, especially for medical conversations and questions.
We believe more efforts should be devoted to the development of new large-scale public biomedical datasets, i.e, unsupervised or distant-supervised automatic annotations and federated learning~\cite{yang2019federated} to allow the development of models while keeping training data on the private side.
Besides, more efforts should be proposed in the future to explore unsupervised, zero-shot, few-shot learning, and data augmentation techniques, with advanced techniques such as LLMs, for low-resource biomedical summarization.

\textbf{Handling long biomedical documents}
PLMs and LLMs generally are limited to a given length of text due to the time complexity of the model.
It is urgent to investigate the effective way for PLMs and LLMs in the biomedical domain to encode the full content of long texts. This could involve investigating segmentation strategies that preserve the semantic integrity of the long biomedical text or adapting PLMs and LLMs to deal with longer contexts without substantial computational costs. The design of domain-specific architecture or techniques like the sparse attention mechanism might offer a path forward.

\textbf{Incorporating extra knowledge} 
PLMs and LLMs are shown to be able to capture common sense knowledge and biomedical knowledge to a certain extent.
Although they can generate summaries that are fluent or grammatically correct, it proves that most of their generated summaries are illogical or have factual errors~\cite{otmakhova2022patient,deyoung2021ms2}.
Limited medical knowledge captured by PLMs and LLMs is hard to support the model to generate desirable summaries.
To overcome this limitation, future research could consider the development of knowledge-aware models, which incorporate additional domain-specific knowledge like UMLS and techniques such as Reinforcement Learning (RL), pre-training, and retrieval augmentation, to enhance the quality and factual correctness of generated summaries. 

\textbf{Controllable generation}
Existing methods generally yield summaries that ignore users' preferences.
We believe more efforts should be developed for controlled summarization of biomedical texts by incorporating PLMs and LLMs with controlled text generation techniques such as prompt engineering, that meet the expectations and requirements of users.
These enhanced methodologies should be designed to manipulate various characteristics of the generated summaries, such as length, readability, and stylistic elements.

\textbf{Benchmarks}
To facilitate the development of biomedical NLP, attempts have been made to create NLP benchmarks in the biomedical domain such as BLUE~\cite{peng2019transfer}, GLUE~\cite{gu2021domain}, which include the relation extraction task, text classification task et al.
However, none of the existing benchmarks includes the biomedical text summarization task.
Biomedical texts encompass a wide array of formats, from scientific papers and Electronic Health Records (EHRs) to dialogues and queries. Furthermore, the task of summarization itself is multifaceted, involving extractive, abstractive, and multi-document summarization. Given this complexity and diversity, we believe it is necessary to build a unified benchmark to support the development and fair evaluations of proposed methods.

\textbf{Multimodality}
The biomedical field boasts a wealth of multimodal medical datasets, such as radiology reports paired with associated X-ray images. Despite this, the majority of existing methods for biomedical text summarization focus solely on the text inputs, overlooking the potential value of associated visual data.
There is increasing evidence that incorporating visual features can significantly enhance text generation performance~\cite{delbrouck2021qiai}. We anticipate a growing interest in multimodal summarization methods in the future that effectively leverage both textual and visual information in the biomedical domain to produce richer, more informative summaries. 

\section{Conclusion}
\label{sec:con}
In this survey, we make a comprehensive overview of biomedical text summarization with pre-trained language models and large language models.
We systematically review recent approaches that are based on PLMs and LLMs, benchmark datasets, evaluations of the task.
We categorize and compare recent approaches according to their utilization of PLMs and LLMs and the specific models they employ.
Finally, we highlight the limitations of existing methods and suggest potential directions for future research.
We hope the paper can be a timely survey to help future researchers.

\section*{Acknowledgement}
This research is partially supported by the Alan Turing Institute and the Biotechnology and Biological Sciences Research Council (BBSRC), BB/P025684/1.
\bibliographystyle{ACM-Reference-Format}
\bibliography{main}

\end{document}